\documentclass{article}

 \usepackage[preprint]{neurips_2026}

\usepackage[utf8]{inputenc} % allow utf-8 input
\usepackage[T1]{fontenc}    % use 8-bit T1 fonts
\usepackage{hyperref}       % hyperlinks
\usepackage{url}            % simple URL typesetting
\usepackage{booktabs}       % professional-quality tables
\usepackage{amsfonts}       % blackboard math symbols
\usepackage{nicefrac}       % compact symbols for 1/2, etc.
\usepackage{microtype}      % microtypography
\usepackage{xcolor}         % colors
\usepackage{tabularx}
\usepackage{graphicx}
\usepackage{subcaption}
\usepackage{wrapfig}
\usepackage[most]{tcolorbox}
\usepackage[normalem]{ulem}
\usepackage{courier}
\usepackage{pifont}
\usepackage{fancyvrb}
\tcbuselibrary{breakable}
\usepackage{subcaption} 
\usepackage{wrapfig}
\usepackage{fvextra}
\usepackage[HTML,table,xcdraw, dvipsnames]{xcolor}

\usepackage{booktabs}
\usepackage{colortbl}

\definecolor{cmark}{HTML}{2D6B42}
\definecolor{xmark}{HTML}{8B2E2E}
\newcommand{\cmark}{\textcolor{cmark}{\textbf{$\checkmark$}}}
\newcommand{\xmark}{\textcolor{xmark}{\textbf{$\times$}}}

\definecolor{ideacol}{RGB}{170,210,202}    % deeper pale teal
\definecolor{codecol}{RGB}{180,212,176}    % deeper pale sage green
\definecolor{writecol}{RGB}{238,200,160}   % deeper pale peach
\definecolor{reviewcol}{RGB}{225,180,170}  % deeper pale dusty rose
\definecolor{litcol}{RGB}{200,180,215}     % deeper pale lavender

\newcommand{\tooltag}[2]{\fcolorbox{#1!60}{#1!20}{\scriptsize\textbf{#2}}}

\title{MLReplicate: Benchmarking Autonomous Research Systems for Machine Learning Reproducibility}

\author{%
  Sasi Kiran Gaddipati\thanks{Equal contribution} $^1$, 
  Diyana Muhammed\footnotemark[1] $^1$, 
  Farhana Keya\footnotemark[1] $^1$,
  Gollam Rabby$^2$\thanks{Equal lead} \thanks{Equal supervision},
  Sören Auer \footnotemark $^1$\\
  $^1$TIB—Leibniz Information Centre for Science and Technology, Hannover, Germany \\
  $^2$L3S Research Center, Leibniz University Hannover, Hannover, Germany \\
  \texttt{Corresponding author: gollam.rabby@l3s.de}
}

\begin{document}

\maketitle

\begin{abstract}
Autonomous research systems capable of generating complete scientific manuscripts have advanced rapidly, yet robust and realistic evaluation frameworks have failed to keep pace. To bridge this gap, we introduce \textbf{MLReplicate}\footnote{Code and data are available:\url{https://github.com/gsasikiran/MLReplicate-benchmarking}}, an end-to-end benchmark evaluating autonomous research systems on machine learning reproducibility. The benchmark was constructed from ICML~2025 outstanding papers reformulated into standardized input specifications and evaluated across 6 state-of-the-art research systems: \textsc{AI Scientist-v1}, \textsc{AI Scientist-v2}, \textsc{Agent Laboratory}, \textsc{CycleResearcher}, \textsc{AI Researcher}, and \textsc{Tiny Scientist}, yielding 45 generated manuscripts, with 3 failed experiments. Outputs are assessed using a dual-protocol approach that combines automated conference-style review and structured expert human evaluation, while tracking computational cost, runtime, and the amount of required human intervention. The automated conference-style review accepted 10 out of 37 valid submissions. An additional 8 submissions were desk-rejected before review for failing to meet the minimum page threshold. In contrast to automated reviews, human reviewers consistently identified methodological flaws, hallucinated experimental results, and reproducibility failures across all systems, and 59\% of accepted automated reviews contained fabricated or unsupported claims. We further find that neither token budget nor computational cost predicts output quality: the cheapest system outperforms the most resource-intensive system in human evaluation, despite a 38-fold difference in input tokens. We thus demonstrate that autonomous research workflow design matters more than the scale of compute. MLReplicate exposes a substantial gap between current autonomous research systems and genuine scientific rigor, and establishes a practical, extensible evaluation framework for systematic progress toward trustworthy AI-driven scientific discovery.
\end{abstract}

\begin{figure}[h]
    \centering
    \includegraphics[width=\columnwidth]{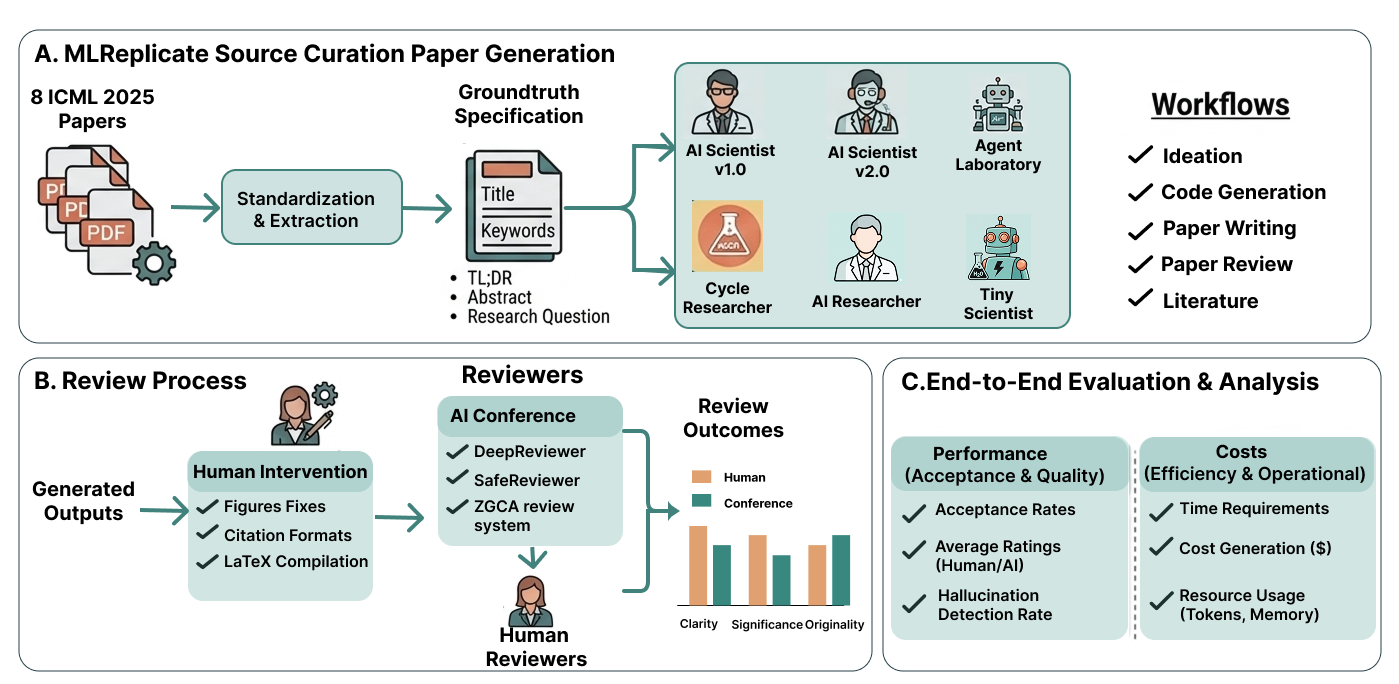}
    \label{fig:overview}
    \caption{\textbf{MLReplicate evaluation pipeline.}
    \textit{(A) Source Curation \& Paper Generation:} 8 ICML 2025 papers are standardized into groundtruth specifications; 6 autonomous systems execute end-to-end research cycles, ideation, code generation, writing, and internal review, to produce generated manuscripts.
    \textit{(B) Review Process:} Outputs undergo human intervention (formatting, citation alignment, \LaTeX{} compilation) before evaluation by both human reviewers and AI reviewers (DeepReviewer, SafeReviewer, ZGCA) across clarity, significance, and originality. \textit{(C) End-to-End Evaluation \& Analysis:} Systems are benchmarked on performance (acceptance rates, ratings, hallucination detection) and efficiency (cost, runtime, resource usage).}
    \label{fig:overview}
    \vspace{-0.3cm}
\end{figure}

\section{Introduction}
Artificial intelligence holds great potential to revolutionize the scientific process, from hypothesis generation~\citep{rabby2025iterative, guo2025ideabench} to automating full experiments~\citep{lu2024ai}. Recent advances in LLMs~\citep{brown2020language, achiam2023gpt, team2023gemini} have enabled autonomous research systems such as \textsc{AI Scientist}~\citep{lu2024ai, yamada2025ai}, \textsc{Agent Laboratory}~\citep{schmidgall2025agent}, and \textsc{Tiny Scientist}~\citep{yu2025tinyscientist} that generate complete manuscripts from a problem specification alone. Yet their ability to produce genuinely valid scientific contributions remains deeply uncertain, due to two key evaluation challenges.

\textbf{First, end-to-end evaluation remains an open problem.} Existing benchmarks address only isolated stages idea generation~\citep{keya2026sci, qiu2025ai}, experimental execution~\citep{chan2024mle, huang2023mlagentbench}, or code reproduction~\citep{starace2025paperbench, xiang2025scireplicate} rather than the full pipeline. End-to-end efforts like MLR-Bench~\citep{chen2025mlr} rely on LLM judges whose alignment with human judgment remains poorly understood~\citep{idahl2025openreviewer, zhu2025deepreview}, making it impossible to distinguish genuine contributions from plausible but superficial outputs.

\textbf{Second, research-level evaluation must account for the practical cost of deployment.} Autonomous research systems differ substantially in computational cost, runtime, and required human intervention, factors that directly determine real-world feasibility, yet virtually all existing benchmarks evaluate quality in isolation from efficiency. A system that produces marginally better outputs at ten times the cost and with extensive human correction may be far less useful in practice than a more efficient alternative. A comprehensive evaluation framework must jointly assess scientific quality and operational cost.

\begin{table*}[t]
\centering
\scriptsize
\setlength{\tabcolsep}{5pt}
\caption{\textbf{Comparison of benchmarks for autonomous research systems.} Column abbreviations: Idea = Ideation, Exp. = Experiment, Imp. = Implementation, Write = Writing, Hum. = Human Review, LLM. = LLM Judge, Int. = Intervention.}
\label{tab:benchmarks}
\begin{tabular}{@{} l *{4}{c} c *{5}{c} @{}}
\toprule
& \textbf{Idea} & \textbf{Exp.} & \textbf{Imp.} & \textbf{Write}
& & \textbf{Hum.} & \textbf{LLM.} & \textbf{Cost} & \textbf{Time} & \textbf{Int.} \\
\midrule
AI Idea Bench \citep{qiu2025ai}    & \cmark & \xmark & \xmark & \xmark && \xmark & \cmark & \xmark & \xmark & \xmark \\
Idea2Plan Bench \cite{huang2025idea2plan}  & \cmark & \xmark & \xmark & \xmark && \xmark & \cmark & \xmark & \xmark & \xmark \\
MLE-Bench \citep{chan2024mle}        & \xmark & \cmark & \xmark & \xmark && \xmark & \xmark & \xmark & \xmark & \xmark \\
MLAgentBench \citep{huang2023mlagentbench}   & \xmark & \cmark & \xmark & \xmark && \xmark & \xmark & \xmark & \xmark & \xmark \\
SUPER \citep{bogin2024super}         & \xmark & \cmark & \cmark & \xmark && \xmark & \xmark & \xmark & \xmark & \xmark \\
CORE-Bench   \citep{siegel2024core}     & \xmark & \cmark & \cmark & \xmark && \xmark & \xmark & \cmark & \xmark & \xmark \\
ScienceAgentBench \citep{chen2024scienceagentbench} & \xmark & \cmark & \cmark & \xmark && \cmark & \cmark & \cmark & \xmark & \xmark \\
ResearchCodeBench \citep{hua2025researchcodebench} & \xmark & \xmark & \cmark & \xmark && \xmark & \xmark & \xmark & \xmark & \xmark \\
SciReplicate-Bench \citep{xiang2025scireplicate} & \xmark & \xmark & \cmark & \xmark && \xmark & \cmark & \xmark & \xmark & \xmark \\
PaperBench   \citep{starace2025paperbench}     & \xmark & \cmark & \cmark & \xmark && \cmark & \cmark & \xmark & \xmark & \xmark \\
\midrule
Scientist-Bench \citep{tang2025ai}  & \cmark & \cmark & \cmark & \cmark && \xmark & \cmark & \xmark & \xmark & \xmark \\
MLR-Bench \citep{chen2025mlr}        & \cmark & \cmark & \cmark & \cmark && \cmark & \cmark & \cmark & \xmark & \xmark \\
\midrule
\textbf{\textsc{MLReplicate} (ours)} & \cmark & \cmark & \cmark & \cmark && \cmark & \cmark & \cmark & \cmark & \cmark \\
\bottomrule
\end{tabular}
\vspace{-0.3cm}
\end{table*}

To address both challenges, we introduce \textsc{MLReplicate}, a benchmark for evaluating autonomous research systems in a standardized end-to-end setting, constructed from 8 ICML 2025 outstanding papers spanning diverse machine learning subfields, including deep learning, probabilistic modeling, generative models, fairness, and meta-science as shown in Figure \ref{fig:overview}. Detailed design features are provided in Appendix~\ref{sec:Features_of_MLReplicate}. Six autonomous research systems are evaluated, producing 45 generated manuscripts assessed through a dual evaluation protocol: automated conference-style review followed by structured expert human evaluation, with full tracking of computational cost, runtime, and human intervention. \autoref{tab:benchmarks} situates \textsc{MLReplicate} as the only framework to jointly evaluate multiple autonomous research systems with multiple evaluation criteria within a single unified framework.

% Our evaluation reveals several striking findings. Out of 48 targeted submissions, 3 failed to generate any output after exhausting the maximum retry limit, 8 were excluded for failing to meet the minimum page requirement (three pages), and only 10 were accepted by the automated review pipeline, proceeding to human evaluation. Human reviewers were significantly more critical, assigning predominantly borderline-reject to reject scores across all systems. Moreover, they identified fabricated or unsupported claims in 59\% of the automated acceptances. The overall correlation between automated and human judgments was weak (Pearson $r = +0.29$), although moderate-to-strong correlations emerged for soundness ($r = +0.77$) and contribution significance ($r = +0.84$). Notably, the cheapest and fastest system outperformed the most resource-intensive one in human evaluation, despite a 38 times difference in input tokens. These results highlight that pipeline design matters far more than raw compute scale.
Our evaluation reveals several striking findings. Out of 48 targeted submissions, 3 failed to generate any output after exhausting the maximum retry limit, 8 were excluded for failing to meet the minimum page requirement (at least 3 pages), and only 10 were accepted by the automated review pipeline and proceeded to human evaluation. Human reviewers were significantly more critical, assigning predominantly borderline-reject to reject scores across all systems. Moreover, they identified fabricated or unsupported claims in 59\% of the automated acceptances. The correlation between automated and human judgments for overall rating is weak (Pearson $r=+0.29$), although moderate-to-strong correlations emerged for soundness ($r=+0.77$) and contribution significance ($r=+0.84$). Notably, the cheapest and fastest system outperformed the most resource-intensive one in human evaluation, despite a 38 times difference in input tokens, highlighting that pipeline design matters far more than raw compute scale. Our contributions include:

\begin{itemize}
    \item \textit{MLReplicate}, a benchmark for end-to-end evaluation of autonomous research systems, constructed from 8 ICML 2025 outstanding papers and 6 state-of-the-art system architectures, yielding 45 generated manuscripts for assessment.
 
    \item A \textit{standardized dual evaluation framework} combining automated conference-style review with structured expert human assessment, enabling systematic analysis of agreement and divergence between automated and human judgment of AI-generated research.
 
    \item A \textit{systematic operational analysis} measuring computational cost, runtime, resource usage, and required human intervention across all evaluated systems, dimensions existing benchmarks overlook but that are essential for real-world deployment decisions.
 
    \item An \textit{empirical characterization} of autonomous research system limitations, including pervasive hallucination, methodological inconsistency, and the absence of a reliable relationship between compute investment and output quality.
\end{itemize}

\section{Experiments: Do they replicate similarly to the human researchers?}
\label{sec:experiments}
This section describes the dataset construction, system interfacing, and evaluation protocol used to assess these autonomous research systems under standardized and reproducible conditions.

\begin{table*}[t]
\centering
\scriptsize
\caption{\textbf{Overview of research papers.} ICML 2025 Outstanding Papers are included in MLReplicate, with their primary topic.}
\label{tab:dataset}
\renewcommand{\arraystretch}{1.1}
\setlength{\tabcolsep}{5pt}
\begin{tabularx}{\textwidth}{p{0.2cm} p{10.3cm} p{3.5cm}}
\toprule
\textbf{ID} & \textbf{Paper Title} & \textbf{Primary Topic} \\
\midrule
P1 & Roll the dice \& look before you leap: Going beyond the creative limits of next-token prediction \citep{nagarajan2025roll}
& Language Models\\
 
P2 & Conformal Prediction as Bayesian Quadrature \citep{snell2025conformal}
& Probabilistic Methods \\
 
P3 & Train for the Worst, Plan for the Best: Understanding Token Ordering in Masked Diffusions \citep{kim2025train}
& Generative Models \\
 
P4 & The Value of Prediction in Identifying the Worst-Off \citep{fischer2025value}
& Fairness \\
 
P5 & CollabLLM: From Passive Responders to Active Collaborators \citep{wu2025collabllm}
& Human-AI Collaboration\\
 
P6 & Score Matching with Missing Data \citep{givens2025score}
& Unsupervised Learning\\
 
P7 & Position: The AI Conference Peer Review Crisis Demands Author Feedback and Reviewer Rewards \citep{kim2025position}
& Meta-Science \\
 
P8 & \textit{Position: AI Safety should prioritize the Future of Work} \citep{hazra2025position}
& AI Safety \\
\bottomrule
\end{tabularx}
\vspace{-0.3cm}
\end{table*}

\paragraph{Experimental Dataset.}
 
The MLReplicate dataset covers diverse machine learning subfields: deep learning, probabilistic modeling, generative models, fairness, and meta-science, encompassing theoretical, empirical, and position papers to ensure varied research styles and methodological complexity, as shown in \autoref{tab:dataset}. Each paper is converted into a standardized ground-truth specification, abstracting the original manuscript into a set of hypotheses, experimental design, evaluation metrics, and key findings, serving as the canonical input for all downstream systems. The dataset is intentionally limited in scale, curated to include only prominent and impactful papers with no overlap with the evaluated LLMs' training data, prioritizing careful curation and rapid iteration over volume: following the philosophy of PaperBench~\citep{starace2025paperbench}, CORE-Bench~\citep{siegel2024core}, and ResearchCodeBench~\citep{hua2025researchcodebench}, which demonstrated that small, rigorous evaluation sets yield reliable and actionable insights. Machine learning constitutes a particularly demanding testbed given its technical complexity, diversity of abstractions, and rapid methodological pace, requiring systems to reason about novel algorithms and non-trivial implementation details. We view MLReplicate as a growing platform, with plans to expand into biology, physics, and materials science, and welcome integration with community-driven efforts such as MLR-Bench~\citep{chen2025mlr} and MLGym~\citep{nathani2025mlgym} to provide an increasingly holistic picture of progress in AI-driven scientific discovery.

\paragraph{Task Construction and System Adaptations.}
% Since autonomous research systems differ substantially in their input requirements and output formats, we design a unified transformation pipeline that converts each standardized ground-truth specification into system specific inputs, adapting structured fields to each system's expected interface such as structured JSON, free form prompts, or notebook templates while collecting outputs in their native formats, typically \LaTeX{} manuscripts or compiled PDFs. Although all systems are used in their publicly released configurations, minor interface-level adaptations are necessary for benchmark compatibility, including mapping standardized specifications to system-specific schemas, applying conference submission templates where required, resolving missing dependencies or configuration mismatches, and ensuring \texttt{.bib} files or textual citations are correctly passed to systems that depend on them. Crucially, none of these changes affect how any system reasons or generates content; all comparisons reflect inherent system capabilities rather than implementation artifacts. Table~\ref{tab:systems_overview} and Appendix~\ref{sec:Benchmark_Metadata_Format} provide a full account of input and output specifications across all evaluated systems.
Since autonomous research systems differ substantially in their input requirements and output formats, we design a unified transformation pipeline that converts each standardized ground-truth specification into system-specific inputs adapting structured fields to each system's expected interface (structured JSON, free-form prompts, or notebook templates) while collecting outputs in their native formats (typically \LaTeX{} manuscripts or compiled PDFs). Although all systems are used in their publicly released configurations, minor interface-level adaptations are necessary for benchmark compatibility, including schema mapping, conference template application, dependency resolution, and correct propagation of \texttt{.bib} files. Crucially, none of these changes affect how any system reasons or generates content; all comparisons reflect inherent system capabilities rather than implementation artifacts. \autoref{tab:systems_overview} provides the full input/output list, while \autoref{sec:Benchmark_Metadata_Format} details the input specification across all evaluated systems.

\paragraph{Dealing with Incomplete Specifications.}
Standardized specifications capture key components of original papers, but inevitably omit implicit design choices and implementation details. 
We address this via four principles: 
\textit{(i) minimal completion}: systems infer missing elements (\emph{e.g.}, hyperparameters, dataset splits) during generation; 
\textit{(ii) no external augmentation}: no information beyond the specification is provided, ensuring identical constraints across systems; 
\textit{(iii) faithfulness over completeness}: the goal is a coherent, plausible research artifact from partial information, not perfect reconstruction; and 
\textit{(iv) consistent input conditions}: all systems receive identical semantic content, isolating system design from input variability. 
This mirrors real-world research settings 
where incomplete information and implicit assumptions are the norm.

\begin{table*}[t]
\centering
\scriptsize
\caption{\textbf{Overview of evaluated automated research systems.} Exp. indicates whether experiments are executed. Tool set badges:
\tooltag{ideacol}{I}~Ideation,
\tooltag{codecol}{C}~Code,
\tooltag{writecol}{W}~Writing,
\tooltag{reviewcol}{R}~Paper Review,
\tooltag{litcol}{L}~Literature Search.}
\label{tab:systems_overview}
\renewcommand{\arraystretch}{1.5}
\setlength{\tabcolsep}{6pt}
\begin{tabularx}{\textwidth}{p{2.5cm} p{1.8cm} p{1.8cm} p{0.2cm} p{2.5cm} p{2.5cm}}
\toprule
\textbf{System} & \multicolumn{1}{c}{\textbf{Inputs}} & \multicolumn{1}{c}{\textbf{Outputs}} & \multicolumn{1}{c}{\textbf{Exp.}} & \multicolumn{1}{c}{\textbf{Tool Sets}} & \multicolumn{1}{c}{\textbf{Models}} \\
\midrule

\textsc{AI Scientist v1}
& \begin{tabular}[t]{@{}l@{}}Template \\[-5pt] Seed ideas\end{tabular}
& \begin{tabular}[t]{@{}l@{}}\LaTeX{}, PDF, Code\end{tabular}
& \cmark & \tooltag{ideacol}{I} \tooltag{codecol}{C} \tooltag{writecol}{W} \tooltag{reviewcol}{R} \tooltag{litcol}{L}
& \begin{tabular}[t]{@{}l@{}}gpt-4o \end{tabular} \\

\textsc{AI Scientist v2}
& \begin{tabular}[t]{@{}l@{}}Title \\[-5pt] Keywords \\[-5pt] Abstract\end{tabular}
& \begin{tabular}[t]{@{}l@{}}\LaTeX{}, PDF\end{tabular}
& \cmark & \tooltag{ideacol}{I} \tooltag{writecol}{W} \tooltag{litcol}{L}
& \begin{tabular}[t]{@{}l@{}}gpt-4o-mini, gpt-4o, o3-mini\end{tabular} \\

\textsc{Agent Laboratory}
& \begin{tabular}[t]{@{}l@{}}Research topic \\[-5pt] Task notes\end{tabular}
& \begin{tabular}[t]{@{}l@{}}\LaTeX{}, PDF, Code\end{tabular}
& \cmark & \tooltag{ideacol}{I} \tooltag{codecol}{C} \tooltag{writecol}{W} \tooltag{reviewcol}{R} \tooltag{litcol}{L}
& \begin{tabular}[t]{@{}l@{}}o3-mini\end{tabular} \\

\textsc{AI Researcher}
& \begin{tabular}[t]{@{}l@{}}Research question \\[-5pt] References\end{tabular}
& \begin{tabular}[t]{@{}l@{}}\LaTeX{}, PDF, Code\end{tabular}
& \cmark & \tooltag{ideacol}{I}\tooltag{codecol}{C} \tooltag{writecol}{W}  \tooltag{litcol}{L}
& \begin{tabular}[t]{@{}l@{}}gpt-4o, gpt-4o-mini\end{tabular} \\

\textsc{CycleResearcher}
& \begin{tabular}[t]{@{}l@{}}Topic \\[-5pt] Bibliography\end{tabular}
& \begin{tabular}[t]{@{}l@{}}\LaTeX{}\end{tabular}
& \xmark 
& \tooltag{writecol}{W}
& \begin{tabular}[t]{@{}l@{}}cycle-researcher-12B\end{tabular} \\

\textsc{Tiny Scientist}
&  Research intent
& \begin{tabular}[t]{@{}l@{}}\LaTeX{}, PDF, Code\end{tabular}
& \cmark & \tooltag{ideacol}{I} \tooltag{codecol}{C} \tooltag{writecol}{W} \tooltag{reviewcol}{R} \tooltag{litcol}{L}
& \begin{tabular}[t]{@{}l@{}}gpt-4o\end{tabular} \\

\bottomrule
\end{tabularx}
\vspace{-.5cm}
\end{table*}

\paragraph{Large Language Models (LLMs).} All evaluated systems rely on LLMs as their core reasoning and generation engines. We use the default or recommended LLM for each system, wherever possible, with all configurations
summarized in \autoref{tab:systems_overview} and further detailed in \autoref{tab:llm_configs} in \autoref{sec:llm_configs}. Across systems, LLM selection introduces an additional axis of variation beyond system architecture; some frameworks rely on a single
general-purpose model, while others decompose tasks across multiple specialized models, complicating direct comparison, as performance differences may stem from both system design and underlying model capability.

\paragraph{Metadata and Reproducibility.}

Additional details on input specifications, output formats, YAML configuration templates, LLM configurations, and evaluation setup are provided in \autoref{sec:Benchmark_Metadata_Format} to support reproducibility.

% \begin{table}[h]
% \centering
% \caption{LLM configurations across evaluated systems. \textbf{Bold} {denotes the
% best-performing or primary model used for the main benchmark results.} \textcolor{red}{Please check this table and fix the information. Also, if you can change the format of this table}}
% \label{tab:llm_configs}
% \renewcommand{\arraystretch}{1.15}
% \setlength{\tabcolsep}{4pt}
% \begin{tabularx}{\columnwidth}{p{3cm} p{7.2cm} p{4cm}}
% \toprule
% \textbf{System} & \textbf{Models Evaluated} & \textbf{Best / Primary} \\
% \midrule
% AI Scientist v1
% & \texttt{gpt-4o-2024-05-13}, \texttt{claude-3-5-sonnet-20241022}
% & \textbf{Both used} \\
 
% AI Scientist v2
% & \texttt{gpt-4o-2024-05-13} (ideas), \texttt{o1-preview-2024-09-12} (writing),
%   \texttt{gpt-4o-2024-11-20} (review), \texttt{o3-mini-2025-01-31} (plots)
% & \textbf{Multi-model pipeline} \\
 
% Agent Laboratory
% & \texttt{o1-preview}, \texttt{o1-mini}, \texttt{gpt-4o}
% & \textbf{\texttt{o1-preview}} \\
 
% AI Researcher
% & \texttt{gpt-4o}
% & \textbf{\texttt{gpt-4o}} \\
 
% Tiny Scientist
% & \texttt{gpt-4o}, \texttt{gpt-4o-mini}, \texttt{o1-mini}, \texttt{o1-preview},
%   \texttt{gpt-3.5-turbo}, \texttt{claude-3-\{haiku,sonnet\}},
%   \texttt{claude-3.5-sonnet}, \texttt{deepseek-\{chat,reasoner\}},
%   \texttt{gemini-1.5-\{flash,pro\}}, \texttt{llama-3.1-40b}
% & \textbf{\texttt{gpt-4o}} \\
% \bottomrule
% \end{tabularx}
% \end{table}

\begin{figure}[t]
    \centering
    \begin{minipage}[c]{0.45\textwidth}
        \centering
        \renewcommand{\arraystretch}{1.3}
        \setlength{\tabcolsep}{3pt}
        \captionof{table}{\textbf{Human-AI correlation ratings across evaluation dimensions.}}
        \label{tab:human_ai_correlation_results}
        \resizebox{\linewidth}{!}{%
        \begin{tabular}{l c c c}
        \toprule
        \textbf{Dimension Pair} & \textbf{Pearson $r$} & \textbf{Spearman $\rho$} & \textbf{Sig.?} \\
        \midrule
        Overall / Rating            & $+0.29$ & $+0.40$ & No \\
        Quality / Soundness         & $+0.77$ & $+0.80$ & Yes ($p < 0.05$) \\
        Clarity / Presentation      & $+0.39$ & $+0.52$ & No \\
        Significance / Contribution & $+0.84$ & $+0.88$ & Yes ($p < 0.01$) \\
        \bottomrule
        \end{tabular}}
    \end{minipage}
    \hfill
    \begin{minipage}[c]{0.52\textwidth}
        \centering
        \includegraphics[width=\linewidth]{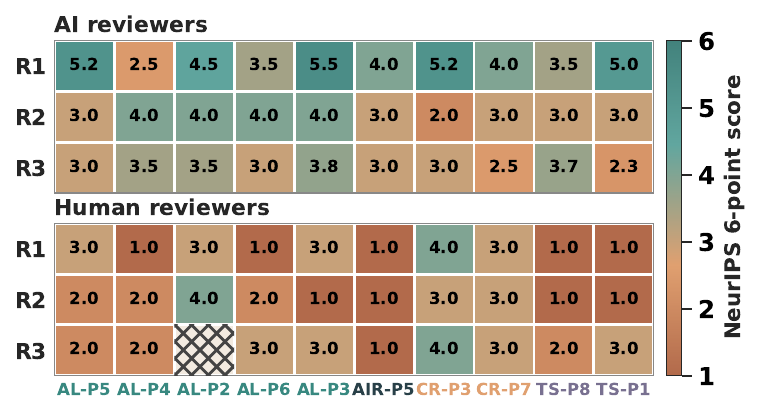}
    \end{minipage}
    \caption{\textbf{Automated vs.\ human evaluation.} Left: correlation analysis between human and AI evaluations. Right: overall rating score distributions across systems. AIR=\textsc{AI Researcher}; AL=\textsc{Agent Laboratory}; CR= \textsc{CycleResearcher}; TS= \textsc{Tiny Scientist}; Pn denotes the paper ID as listed in  Table \ref{tab:dataset}.}
    \label{fig:autovshuman}
    \vspace{-0.7cm}
\end{figure}

\section{Evaluation and Results}
 
 Research quality is assessed through an automated review pipeline followed by expert human evaluation. Operational requirements are quantified through metrics for cost, runtime, and the degree of human intervention. The following sections detail the post-processing applied to system outputs, followed by a progression from automated assessment to human judgment and resource analysis.

 \begin{wrapfigure}{r}{0.5\columnwidth}
    \centering
    % \vspace{-10pt}
    \includegraphics[width=0.40\columnwidth]{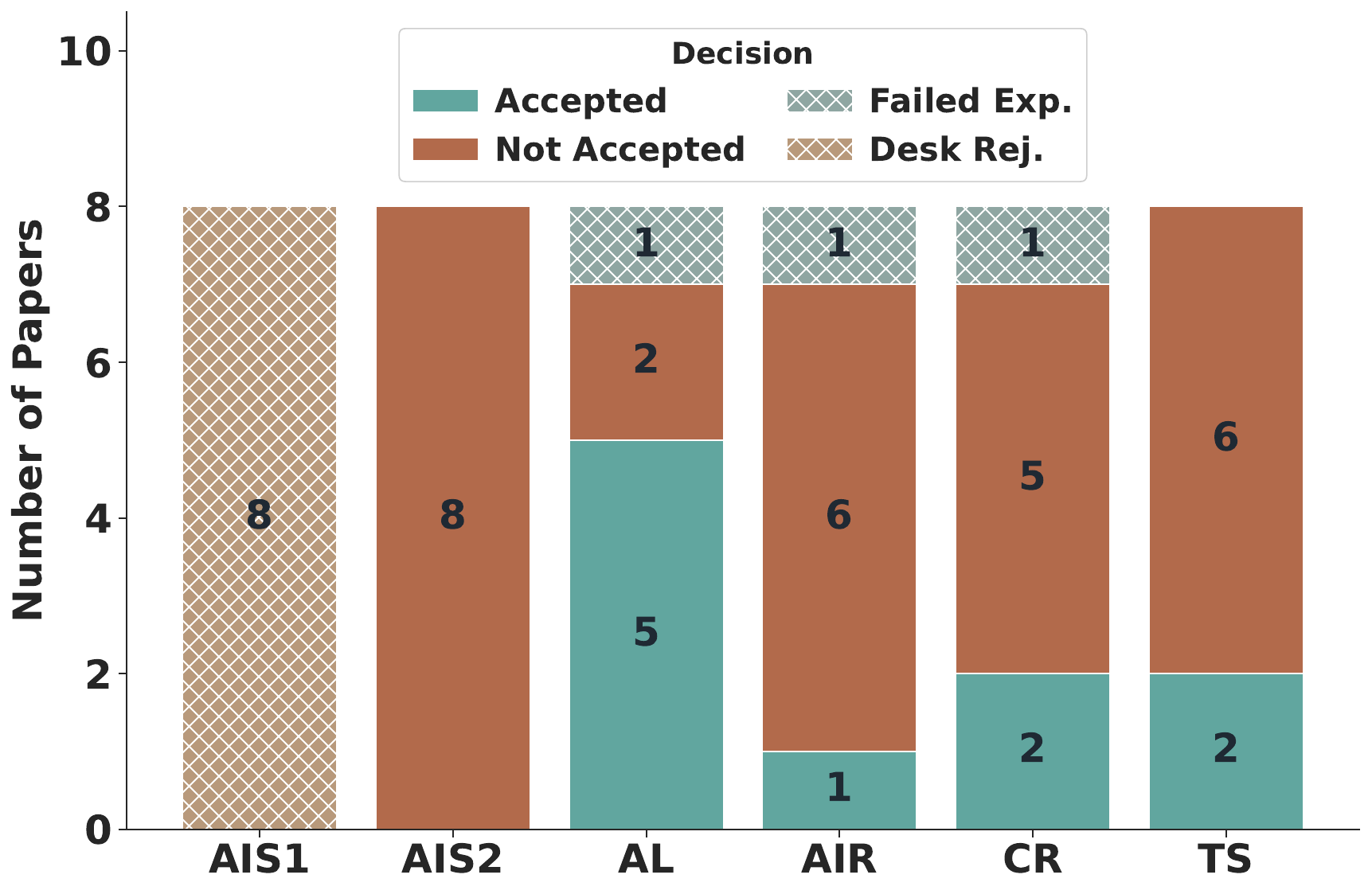}
    % \vspace{-10pt}
    \caption{\textbf{Generated experiments and acceptance distribution for each research system from automated reviews.} AIS1 = \textsc{AI Scientist V1}; AIS2 = \textsc{AI Scientist V2}.}
    \label{fig:icais_conf}
    \vspace{-10pt}
\end{wrapfigure}

\paragraph{Post-Generation Processing.}
\label{sec:postprocessing}

All system outputs undergo limited post-processing before evaluation, strictly confined to formatting corrections, citation alignment, and figure integration fixes; no modifications are made to scientific content, experimental results, or methodological claims. The degree of intervention varies by system: \textsc{CycleResearcher} requires the most correction (formatting inconsistencies, incomplete references, invalid artifacts), followed by \textsc{Agent Laboratory} and \textsc{AI Scientist v1} requiring moderate adjustments (citation formatting, figure integration). \textsc{AI Researcher}, \textsc{Tiny Scientist}, and \textsc{AI Scientist-v2} required only minor fixes regarding \LaTeX{} compilation issues, template and alignment.

% \textcolor{brown}{\textsc{AI Scientist v1} falls into the moderate intervention category, typically requiring corrections to citation consistency, figure placement, and occasional artifact validation, while leaving the underlying scientific content unchanged.}

\paragraph{Automated Review.}
\label{sec:ai_conference_results}

\begin{figure}[t]
    \centering
    \begin{subfigure}{0.45\columnwidth}
        \centering
        \includegraphics[width=\linewidth]{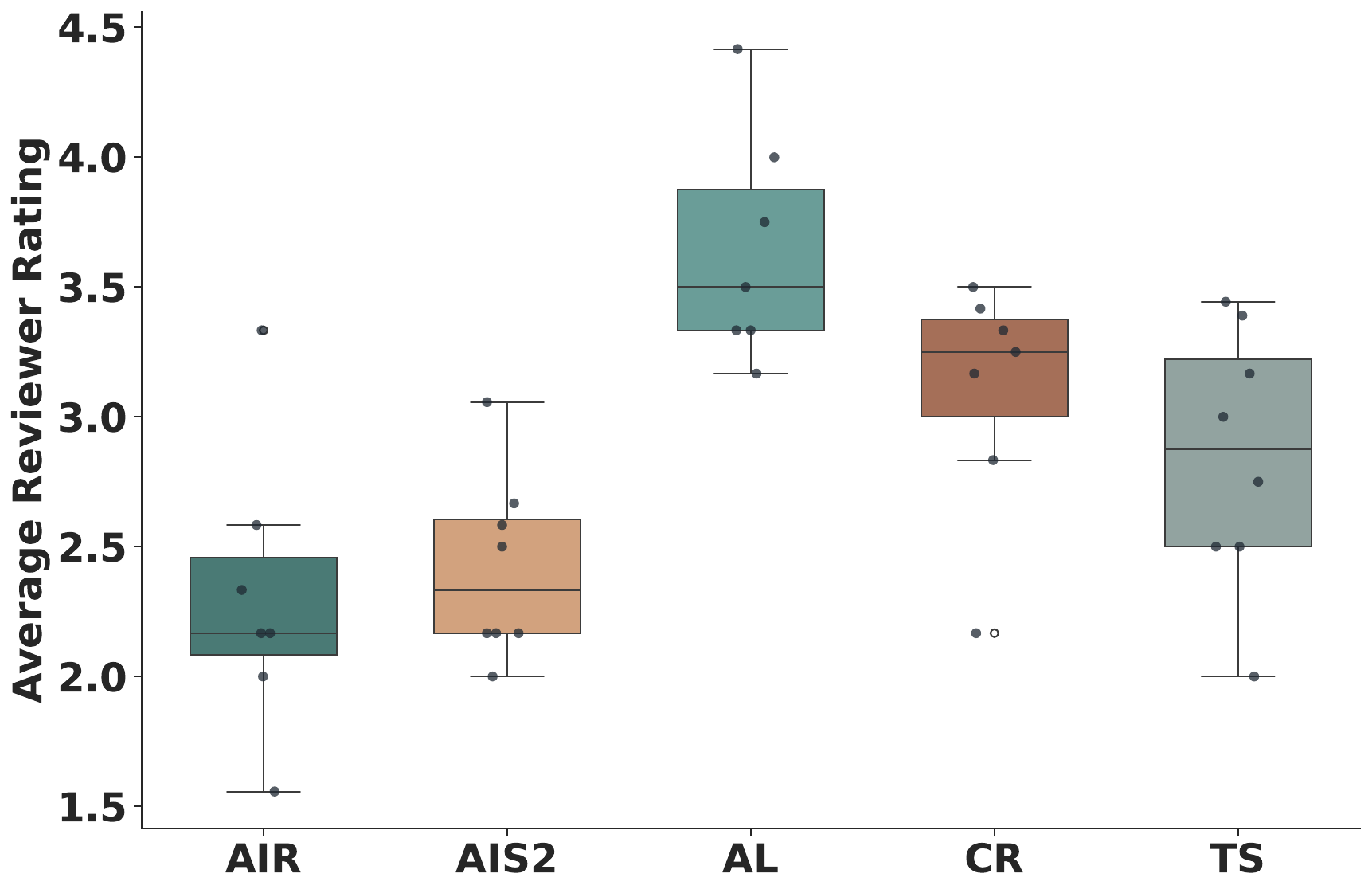}
        \caption{Average reviewer rating distribution (max=6)}
        \label{fig:ai_conf_rating_distribution}
    \end{subfigure}
    \hfill
    \begin{subfigure}{0.45\columnwidth}
        \centering
        \includegraphics[width=\linewidth]{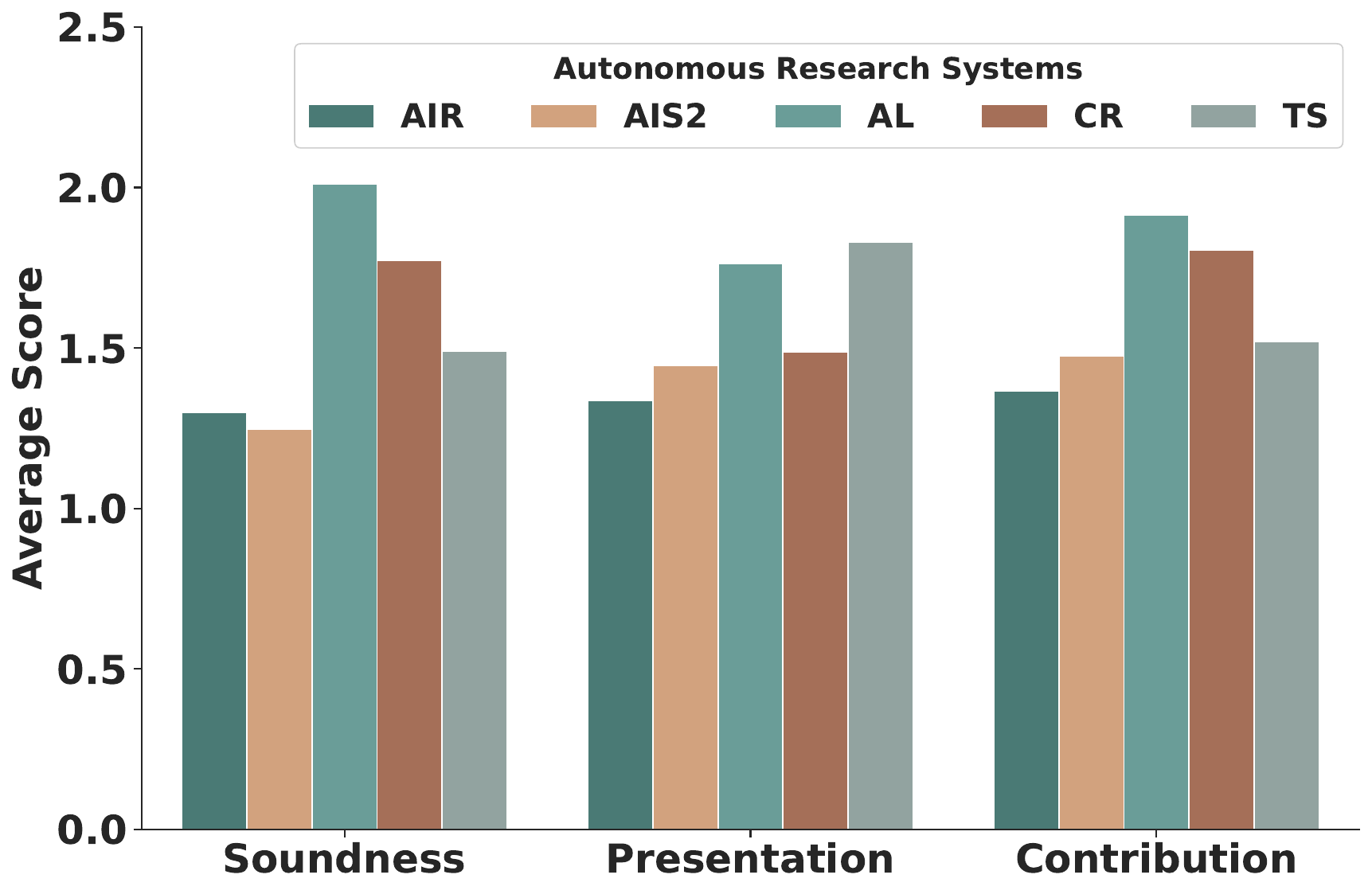}
        \caption{Average dimensional rating (max=4)}
        \label{fig:ai_conf_dimension_evaluation}
    \end{subfigure}

    \caption{\textbf{Overview of overall rating and dimension analysis of the automated reviews.} The short forms represent the relevant research systems. AIR=\textsc{AI Researcher}; AIS2 = \textsc{AI Scientist V2};  AL= \textsc{Agent Laboratory}; CR= \textsc{CycleResearcher}; TS= \textsc{Tiny Scientist}.}
    \label{fig:ai_conference_results}
    \vspace{-0.7cm}
\end{figure}

The 37 processed papers were submitted to the automated review pipeline of the 1st International Conference on AI Scientists (ICAIS 2025), further detailed in Appendix~\ref{sec:ai_conference}, where each submission was independently evaluated by three automated reviewer models: DeepReviewer~\citep{zhu2025deepreview}, the ZGCA review system, and SafeReviewer~\citep{xin2026safereview}, with reviewer scores, written feedback, and final acceptance decisions collected for each paper. As shown in \autoref{fig:icais_conf}, only 10 of the 37 submissions were accepted. \textsc{Agent Laboratory} accounted for 5 of these, the strongest performance among all 6 systems, while \textsc{AI Scientist-v2} received no acceptances, which we attribute primarily to limitations in its text generation quality. No individual paper topic was accepted across more than one system, and only one topic~\citep{wu2025collabllm} was accepted by two systems, reflecting low cross-system consistency on identical inputs. As shown in Figures~\ref{fig:ai_conf_rating_distribution} and~\ref{fig:ai_conf_dimension_evaluation}, \textsc{Agent Laboratory} further achieved the highest median rating (${\approx}3.5$ out of 4) and the top individual score of 4.5, consistently outperforming other systems on soundness, presentation, and contribution. Nevertheless, no system exceeded an average dimensional score of 2 out of 4, underscoring a substantial gap between current autonomous research outputs and the bar set by even an automated review pipeline. Overview of overall rating and dimension analysis at the AI conference shown in Figure \ref{fig:ai_conference_results}.

\begin{figure}[t]
    \centering

     \begin{subfigure}{0.45\columnwidth}
        \centering
        \includegraphics[width=0.90\linewidth]{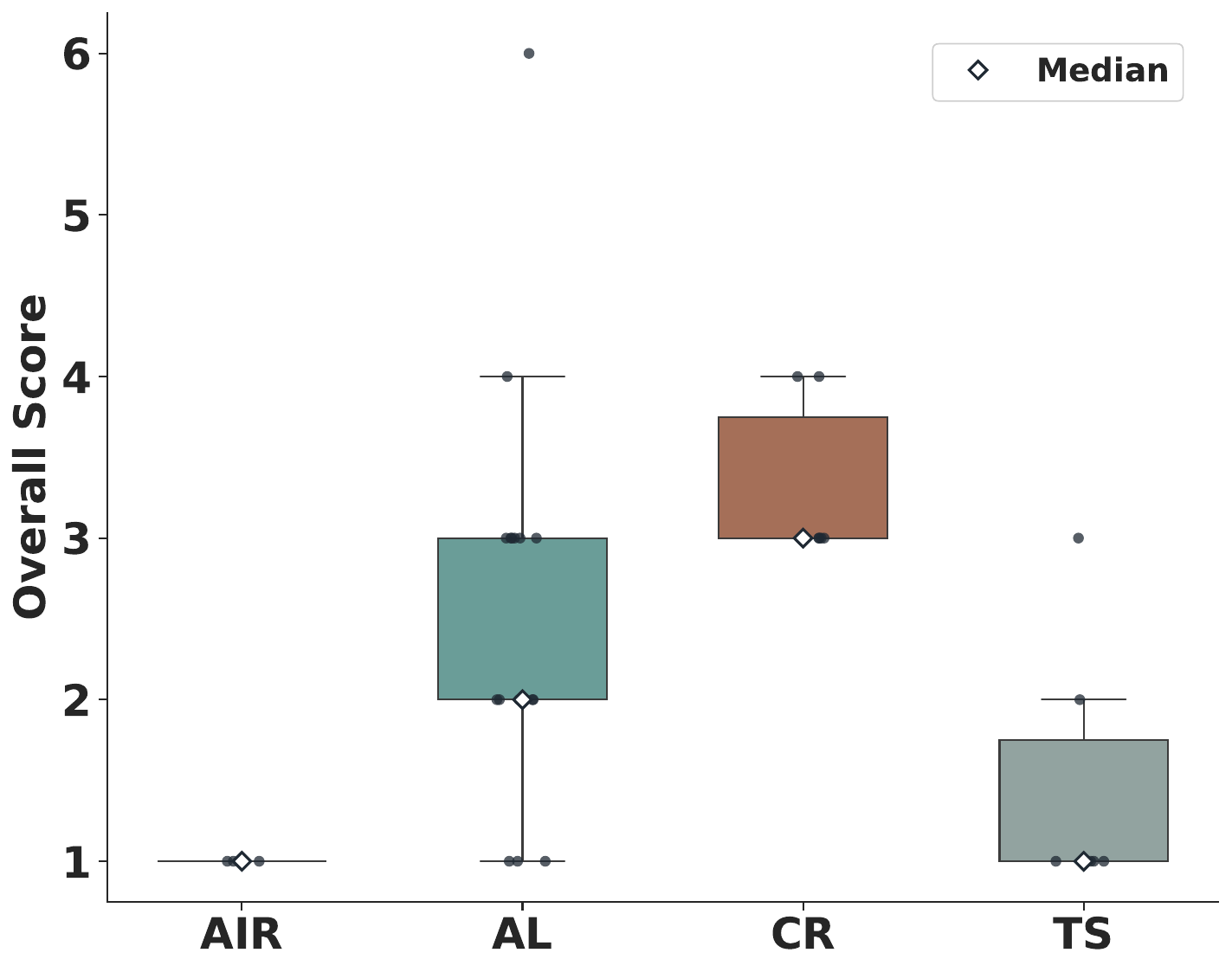}
        \caption{Overall rating distribution (max=6)}
        \label{fig:human_eval_overall_score}
    \end{subfigure}
    \hfill
    \begin{subfigure}{0.45\columnwidth}
        \centering
        \includegraphics[width=0.90\linewidth]{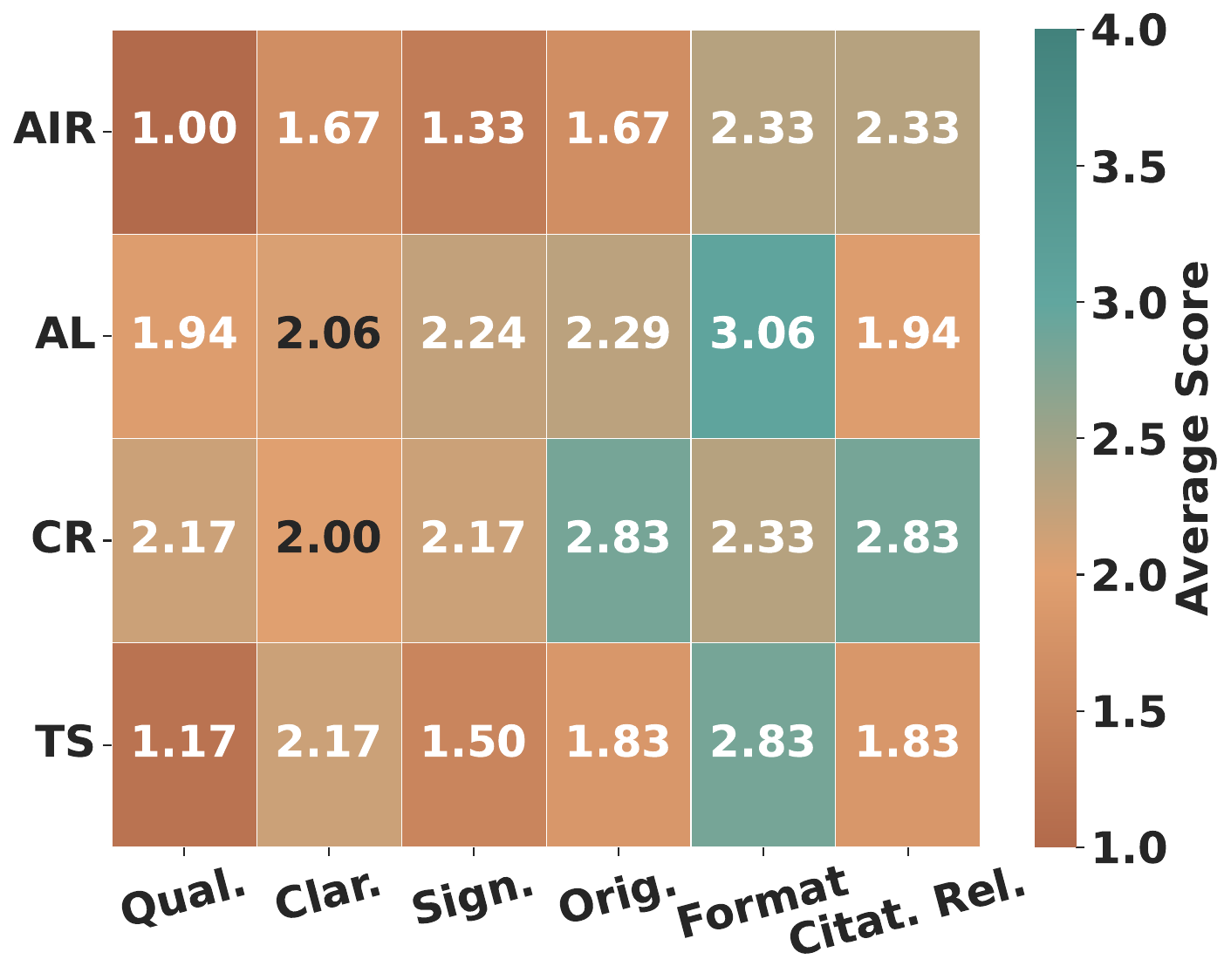}
        \caption{Average dimensional heatmap (max=4)}
        \label{fig:human_eval_dimension_heatmap}
    \end{subfigure}
    
    % \hfill
    % \begin{subfigure}{0.32\columnwidth}
    %     \centering
    %     \includegraphics[width=\linewidth]{img/04_dimension_heatmap.pdf}
    %     \caption{Dimension heatmap of the papers}
    %     \label{fig:dimension_heatmap}
    % \end{subfigure}

    \caption{\textbf{Overview of acceptance statistics and dimension analysis of the human reviewers.} The horizontal axis of the heatmap represents the Quality, Clarity, Significance, Originality, Format, and Citation Relevance in the respective order. AIR=\textsc{AI Researcher}; AL=\textsc{Agent Laboratory}; CR=\textsc{CycleResearcher}; TS=\textsc{Tiny Scientist}.}
    \label{fig:combined_figures}
\end{figure}
\paragraph{Human Expert Evaluation.}
\label{sec:human_eval}

The 10 papers accepted by ICAIS were subsequently evaluated by external human reviewers, each receiving three independent reviews via a standardized form. All reviewers hold a minimum of graduate-level research experience in machine learning, with the majority being active researchers or faculty with prior experience as reviewers at top-tier venues (such as
NeurIPS, ICML, ICLR). Reviewers were explicitly informed that all submissions are AI-generated to prevent unwarranted charitable interpretation. Each submission was assessed across six dimensions on a four-point scale: technical quality, clarity, significance, originality, formatting, and citation relevance, alongside a binary hallucination judgment, free text feedback, and an overall recommendation on a six-point scale from strong reject to strong accept. To further align with NeurIPS review standards, reviewers were additionally asked to assess: \textit{(i)} the soundness and validity of
theoretical claims and proofs, where applicable; \textit{(ii)} the appropriateness and rigor of experimental design, including dataset selection, baseline comparisons, and ablation studies; \textit{(iii)} the presence and adequacy of statistical reporting, such as error bars, confidence intervals, and significance tests; \textit{(iv)} ethical considerations and potential societal impacts of the proposed work; and \textit{(v)} whether the paper's claims are adequately supported by the provided evidence and whether results are likely to be reproducible. Inter-reviewer agreement was monitored to ensure consistency, and cases of high disagreement were flagged for discussion.

We also noticed that human reviewers have been markedly more critical than the automated pipeline. As shown in \autoref{fig:human_eval_overall_score}, average overall ratings fall in the borderline-reject to reject range across all systems: \textsc{Agent Laboratory}, despite its strong automated performance, exhibits high variance in human scores, suggesting inconsistent quality across topics; the single \textsc{AI Researcher} paper was unanimously rated strong reject; and \textsc{CycleResearcher} performed marginally better, with two instances receiving borderline-accept scores from individual reviewers. No system produced outputs that human reviewers considered reliably publication-ready. Hallucination is pervasive: reviewers identified fabricated or unsupported claims in 59\% of all evaluations, with 100\% for \textsc{AI Researcher}, 67\% for \textsc{Tiny Scientist}, 59\% for \textsc{Agent Laboratory}, and 33\% for \textsc{CycleResearcher}. The dimension-level heatmap in \autoref{fig:human_eval_dimension_heatmap} further reveals that \textsc{CycleResearcher} performs relatively better on citation relevance and originality, while \textsc{Agent Laboratory} and \textsc{Tiny Scientist} score higher on formatting; no system performs well across all dimensions simultaneously.

\begin{figure}[t]
    \centering

     \begin{subfigure}{0.45\columnwidth}
        \centering
        \includegraphics[width=0.90\linewidth]{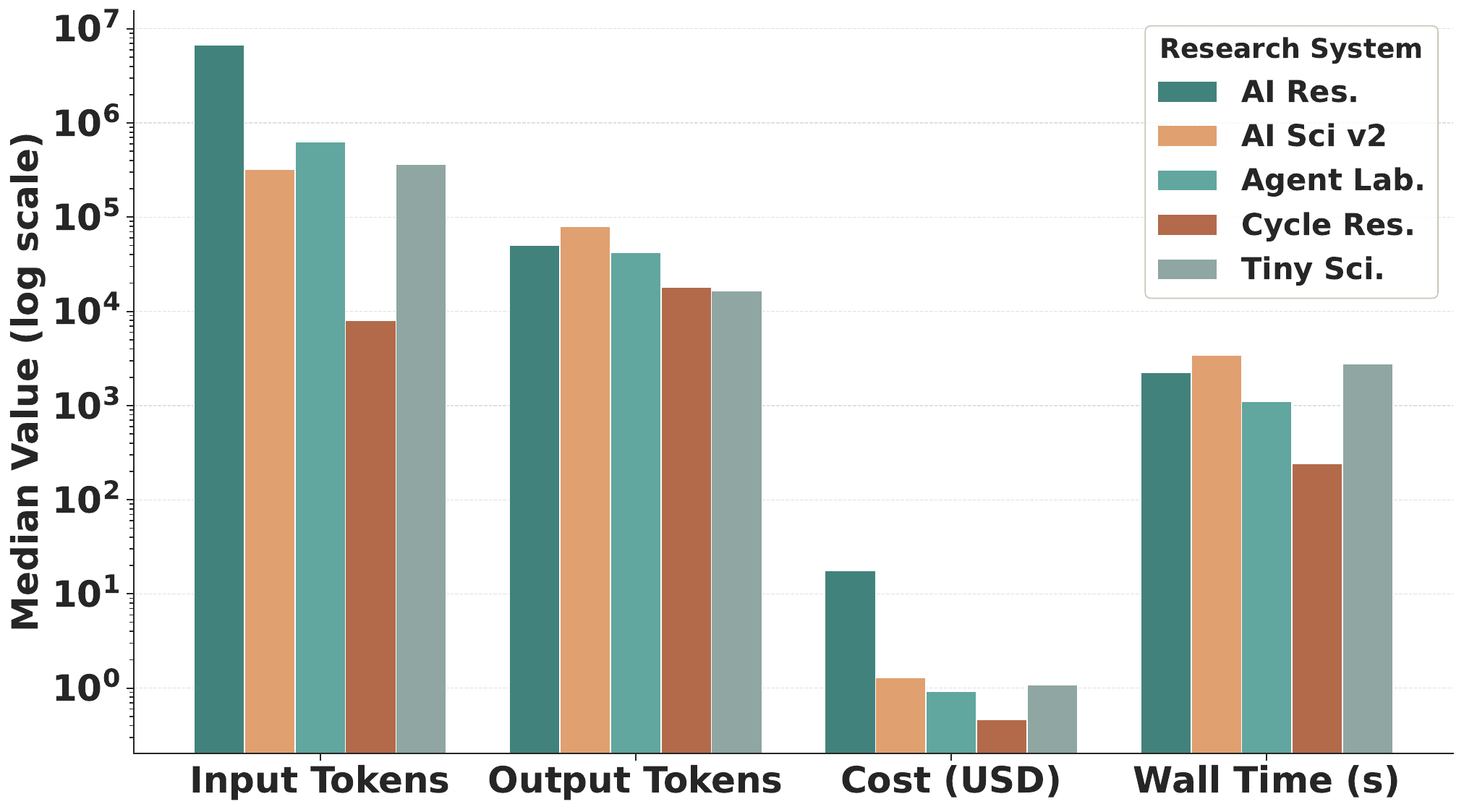}
        \caption{Overall resource distribution }
        \label{fig:resource_distribution}
    \end{subfigure}
    \hfill
    \begin{subfigure}{0.45\columnwidth}
        \centering
        \includegraphics[width=0.90\linewidth]{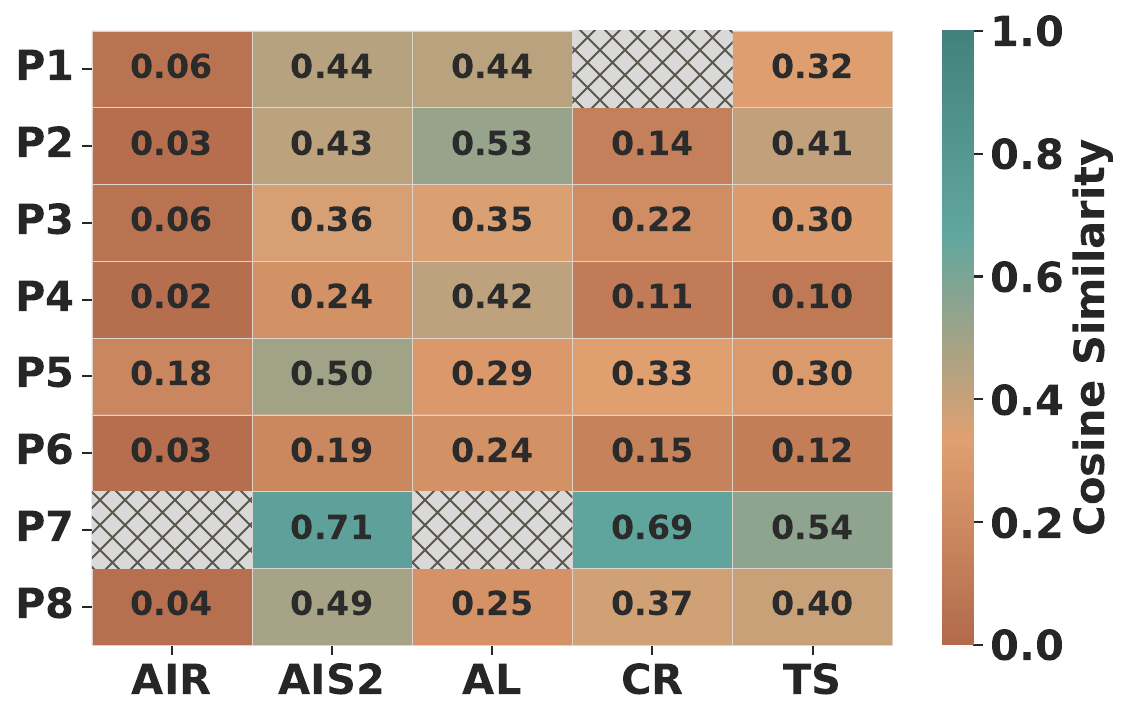}
        \caption{Paper similarity heatmap}
        \label{fig:similarity_heatmap}
    \end{subfigure}
    
    % \hfill
    % \begin{subfigure}{0.32\columnwidth}
    %     \centering
    %     \includegraphics[width=\linewidth]{img/04_dimension_heatmap.pdf}
    %     \caption{Dimension heatmap of the papers}
    %     \label{fig:dimension_heatmap}
    % \end{subfigure}

    \caption{\textbf{Overall resource distribution and a paper similarity heatmap illustrating the similarities between generated and actual papers.} AIS2=\textsc{AI Scientist-v2}, AIR=\textsc{AI Researcher}; AL=\textsc{Agent Laboratory}; CR= \textsc{CycleResearcher}; TS= \textsc{Tiny Scientist}; Pn denotes the paper ID as listed in  Table \ref{tab:dataset}.}
    \label{fig:combined_figures}
    \vspace{-0.7 cm}
\end{figure}

\paragraph{Automated vs. Human Agreement.}
\label{sec:agreement}
 
\autoref{fig:autovshuman} reports the correlation between human reviewer scores and AI conference ratings across the 10 jointly evaluated papers. The correlation between automated and human overall ratings is weak and statistically non-significant (Pearson $r = +0.29$, Spearman $\rho = +0.40$), indicating that automated reviewers do not reliably capture human judgment of overall paper quality. A more nuanced picture emerges at the dimension level: moderate-to-strong, statistically significant correlations are observed for Quality/Soundness ($r = {+}0.77$, $p < 0.05$) and Significance/Contribution ($r = {+}0.84$, $p < 0.01$), while Clarity/Presentation remains non-significant. This suggests that automated reviewers can capture certain underlying evaluation dimensions consistently with human reviewers, but fail to translate these into aligned holistic assessments, indicating that automated review pipelines may be appropriate for coarse-grained filtering, but are not a reliable substitute for expert human judgment on overall scientific quality. This also highlights the need for improved robustness in automated review systems and underscores the importance of further benchmark evaluations tailored to assess their reliability.

\paragraph{Computational Cost and Resource Usage.}
\label{sec:resources}

As illustrated in \autoref{fig:resource_distribution}, resource usage varies significantly across systems and does not correlate directly with output quality. Among the evaluated methods, \textsc{CycleResearcher} appears to be the most efficient in terms of tokens and cost, using 8.1K input tokens and 18.0K output tokens at a cost of \$0.46 with a runtime of 243s. However, it consistently requires 23GB of GPU memory on a single H100. Its cost estimates are based on AWS GPU pricing at the time of evaluation. This efficiency is partly attributable to its limited capabilities, as it does not perform experimentation or literature search, focusing solely on text generation. \textsc{Agent Laboratory} represents a middle ground, consuming 626K input tokens and producing 42.3K output tokens, with a cost of \$0.92 and a runtime of 1,105s. \textsc{Tiny Scientist} exhibits similar intermediate behavior, using 367K input tokens and 16.5K output tokens, costing \$1.09 with a runtime of 2,774s. In contrast, \textsc{AI Researcher} is the most resource-intensive system, requiring 6.80M input tokens and generating 50.3K output tokens, with a median cost of \$17.72 and a runtime of 2,252. \textsc{AI Scientist-v1} serves as a high-cost baseline, taking approximately 28,800s and incurring around \$15.00 in API fees per paper. We exclude resource evaluations for invalidated papers from the figure for clarity. Overall, \textsc{Agent Laboratory} demonstrates a stronger cost-to-performance ratio compared to \textsc{AI Researcher}, which is both more expensive and performs worse in human evaluations. Despite consuming roughly $11\times$ more input tokens than \textsc{Agent Laboratory} on average, \textsc{AI Researcher} does not achieve better results, highlighting that token usage and runtime are poor proxies for output quality and that effective pipeline design is more important than sheer compute scale.

\paragraph{Semantic Similarity Analysis.}
\label{sec:similarity}

To assess output source alignment, we embed all documents in a shared TF-IDF/LSA space and compute pairwise cosine similarities (\autoref{fig:similarity_heatmap}). \textsc{AI Scientist-v2} achieves the strongest alignment (mean 0.42, peak 0.71~\citep{kim2025position}), followed by \textsc{Agent Laboratory} and \textsc{Tiny Scientist} (mean 0.31), \textsc{CycleResearcher} (mean 0.25), and \textsc{AI Researcher} (mean 0.05). Scores correlate poorly with evaluation metrics; the primary driver is input richness where systems receiving abstracts and keywords produce outputs closer to source methodology than those operating from titles or reference lists alone, confirming that semantic similarity is a poor standalone quality metric and that input specification is a consequential, often overlooked variable.

\section{Related Work}

Early applications of artificial intelligence in science focused on augmenting specific components of the research workflow. LLMs were initially used for tasks such as literature understanding, scientific writing, and code generation \citep{wang2023scientific}. While effective in isolated settings, these models lacked the ability to interact with external tools or execute experiments. Subsequent work extended LLMs with domain-specific tools, enabling active participation in scientific workflows. ChemCrow \citep{bran2023chemcrow} integrated chemistry tools to support synthesis planning, while \citet{boiko2023autonomous} demonstrated autonomous experimental design and execution in laboratory environments. These systems showed that LLMs can move beyond passive text generation toward actionable scientific reasoning. However, they remained constrained to narrow domains and did not address the generation of complete research outputs.

\paragraph{Autonomous Research Systems.} More recently, a new class of systems has emerged that attempts to automate the full research lifecycle. \textsc{AI Scientist-v1} \citep{lu2024ai} introduced an end-to-end pipeline that generates research ideas, executes experiments, and writes manuscripts through iterative refinement. This framework established a template that subsequent systems have extended in different directions. Several approaches focus on improving exploration and refinement. \textsc{CycleResearcher} \citep{weng2024cycleresearcher} introduces explicit feedback loops between generation and review, while \textsc{AI Scientist-v2} \citep{yamada2025ai} replaces linear pipelines with tree-based search over research directions, enabling broader hypothesis exploration before commitment. Other systems explore alternative architectural designs and levels of autonomy. \textsc{Agent Laboratory} \citep{schmidgall2025agent} decomposes the workflow into specialized agents with optional human feedback, improving interpretability and control. \textsc{Tiny Scientist} \citep{yu2025tinyscientist} emphasizes modularity and configurability, while \textsc{AI Researcher} \citep{tang2025ai} adopts a unified agentic framework with minimal stage separation. Collectively, these systems represent diverse design choices, but they have not been systematically compared under standardized evaluation conditions.

\paragraph{Evaluation of Autonomous Research Systems.} Evaluating autonomous research systems remains an open problem. Most existing benchmarks assess individual components of the research process rather than full end-to-end generation. IdeaBench \citep{guo2025ideabench} and AI Idea Bench \citep{qiu2025ai} focus on idea quality, while MLE-Bench \citep{chan2024mle}, MLAgentBench \citep{huang2023mlagentbench}, and SUPER \citep{bogin2024super} evaluate experimental execution and machine learning workflows. Code- and reproduction-focused benchmarks such as ResearchCodeBench \citep{hua2025researchcodebench}, SciReplicate-Bench \citep{xiang2025scireplicate}, and PaperBench \citep{starace2025paperbench} assess implementation fidelity. However, performance on individual stages does not necessarily reflect the ability to generate a complete scientific contribution. End-to-end benchmarks such as MLR-Bench \citep{chen2025mlr} and MLGym \citep{nathani2025mlgym} evaluate full research outputs using LLM-based judges. However, prior work has shown that automated evaluators can be sensitive to prompting, biased toward surface-level fluency, and misaligned with expert human judgment in scientific evaluation settings \citep{idahl2025openreviewer, zhu2025deepreview, zhu2025ai}. Moreover, these benchmarks typically ignore practical factors such as computational cost, runtime, and human intervention, limiting their usefulness for comparing real-world systems. MLReplicate addresses these limitations by providing a standardized, end-to-end evaluation framework that combines automated and expert human assessment while systematically tracking efficiency-related metrics.

\section{Limitations and Future Work}
\label{sec: limitations}

While MLReplicate provides a rigorous starting point for end-to-end evaluation of autonomous research systems, several limitations remain. First, MLReplicate is currently restricted to machine learning, and we plan to extend coverage to biology, physics, chemistry, and materials science, where agentic systems face structurally distinct challenges. Second, the dataset is limited to 8 papers to ensure annotation quality, constraining statistical power and generalizability; we plan to scale to larger, more diverse datasets alongside stronger autonomous evaluators. Third, the benchmark evaluates generated manuscripts without validating underlying experiments, as systems such as CycleResearcher may report results without executing actual code; unlike standard benchmarks, scientific validation requires executable and reproducible environments, which we view as an essential direction for future work. Fourth, observed failures, including formatting issues yielding invalid outputs, motivate real-time benchmarking infrastructure with automated validation pipelines and leaderboard systems for continuous, large-scale evaluation. Finally, using AI systems to screen AI-generated research may introduce an echo-chamber risk, where model-specific biases may reinforce recurring patterns; more robust hybrid frameworks combining automated methods with expert human judgment are needed. Nonetheless, MLReplicate's focus on end-to-end scientific validity, dual evaluation protocol, and cost tracking makes it a strong testbed for autonomous research systems even at its current scale.

\section{Ethical Considerations}
The development of MLReplicate is grounded in established ethical and legal frameworks for text and data mining (TDM) and fair use, specifically optimized for the evaluation of autonomous research systems. Our methodology adheres to the TDM exception for scientific research under Article 3 of the EU DSM Directive (2019/790/EU) and \S 60d of the German UrhG, which permits the non-commercial processing of copyrighted works for scientific purposes. Within the United States jurisdiction, our framework is supported by a four-factor fair use analysis: the purpose is strictly transformative, converting peer-reviewed research into standardized, machine-readable specifications to benchmark AI reasoning; the nature of the source ICML 2025 outstanding papers is technical and factual; we provide only high-level scientific components hypotheses, experimental designs, and evaluation criteria rather than redistributing full manuscripts; and our structured index does not substitute for the original works but serves to facilitate rigorous replication. All papers are credited to their original authors via comprehensive metadata, and to address the risk of AI-driven misinformation, MLReplicate explicitly incorporates expert human evaluation to identify pervasive hallucinations and unsupported claims, which were identified in 59\% of automated acceptances. This approach is intended for research evaluation in controlled settings and must not be used to generate automated research papers without human control or safety-critical literature, with the full dataset and evaluation suite released under a CC-BY-4.0 license.

\section{Conclusion}
We introduced MLReplicate, a rigorously curated benchmark for end-to-end evaluation of autonomous research systems on machine learning reproducibility. Our comprehensive assessment across six state-of-the-art systems reveals substantial and consistent limitations: hallucination and methodological inconsistency are pervasive, citation consistency and reproducibility remain unsolved, and no system produced outputs that human reviewers considered reliably publication-ready. The strongest-performing system, \textsc{Agent Laboratory}, achieved only a single strong-accept rating with an average human score of 3.2 and high cross-topic variance, while \textsc{AI Scientist-v2} failed to pass automated screening for even a single paper. These results expose a clear and significant gap between current claims of conference-level autonomous research generation and the quality that expert human evaluation actually observes. More broadly, our findings demonstrate that neither token budget nor computational cost predicts output quality, and that automated review pipelines, while useful for coarse filtering, are not a reliable substitute for human judgment on scientific validity. We position MLReplicate as a foundational step toward principled, reproducibility-grounded evaluation of automated research systems, and emphasize the urgent need for continued progress in long-horizon reasoning, experimental rigor, and hallucination control to close the gap between autonomous research generation and genuine scientific contribution.

\section{Author Contributions}

Gollam designed the project, led the experimental design, and wrote the final version of the manuscript. Sasi, Diyana, and Farhana performed the experiments, authored the initial manuscript draft, and performed subsequent revisions. Specifically, Sasi executed the Agent Laboratory and Cycle Researcher frameworks; Diyana conducted experiments using AI Researcher and Tiny Scientist; and Farhana managed AI-Scientist-v1 and AI-Scientist-v2. Gollam and Sören provided scientific supervision, technical advice, and critical feedback throughout the duration of the project.

\section{Acknowledgements}

This work was supported by the BMBF (KISSKI project, 01IS22093C) and the DFG (NFDI4DataScience consortium, project 460234259).

\bibliographystyle{ACM-Reference-Format}
\bibliography{References}

@article{achiam2023gpt,
  title={Gpt-4 technical report},
  author={Achiam, Josh and Adler, Steven and Agarwal, Sandhini and Ahmad, Lama and Akkaya, Ilge and Aleman, Florencia Leoni and Almeida, Diogo and Altenschmidt, Janko and Altman, Sam and Anadkat, Shyamal and others},
  journal={arXiv preprint arXiv:2303.08774},
  year={2023}
}

@inproceedings{bogin2024super,
  title={Super: Evaluating agents on setting up and executing tasks from research repositories},
  author={Bogin, Ben and Yang, Kejuan and Gupta, Shashank and Richardson, Kyle and Bransom, Erin and Clark, Peter and Sabharwal, Ashish and Khot, Tushar},
  booktitle={Proceedings of the 2024 Conference on Empirical Methods in Natural Language Processing},
  pages={12622--12645},
  year={2024}
}

@article{brown2020language,
  title={Language models are few-shot learners},
  author={Brown, Tom and Mann, Benjamin and Ryder, Nick and Subbiah, Melanie and Kaplan, Jared D and Dhariwal, Prafulla and Neelakantan, Arvind and Shyam, Pranav and Sastry, Girish and Askell, Amanda and others},
  journal={Advances in neural information processing systems},
  volume={33},
  pages={1877--1901},
  year={2020}
}

@article{chan2024mle,
  title={Mle-bench: Evaluating machine learning agents on machine learning engineering},
  author={Chan, Jun Shern and Chowdhury, Neil and Jaffe, Oliver and Aung, James and Sherburn, Dane and Mays, Evan and Starace, Giulio and Liu, Kevin and Maksin, Leon and Patwardhan, Tejal and others},
  journal={arXiv preprint arXiv:2410.07095},
  year={2024}
}

@article{chen2024scienceagentbench,
  title={Scienceagentbench: Toward rigorous assessment of language agents for data-driven scientific discovery},
  author={Chen, Ziru and Chen, Shijie and Ning, Yuting and Zhang, Qianheng and Wang, Boshi and Yu, Botao and Li, Yifei and Liao, Zeyi and Wei, Chen and Lu, Zitong and others},
  journal={arXiv preprint arXiv:2410.05080},
  year={2024}
}

@article{chen2025mlr,
  title={Mlr-bench: Evaluating ai agents on open-ended machine learning research},
  author={Chen, Hui and Xiong, Miao and Lu, Yujie and Han, Wei and Deng, Ailin and He, Yufei and Wu, Jiaying and Li, Yibo and Liu, Yue and Hooi, Bryan},
  journal={arXiv preprint arXiv:2505.19955},
  year={2025}
}

@article{fischer2025value,
  title={The value of prediction in identifying the worst-off},
  author={Fischer-Abaigar, Unai and Kern, Christoph and Perdomo, Juan Carlos},
  journal={arXiv preprint arXiv:2501.19334},
  year={2025}
}

@article{givens2025score,
  title={Score matching with missing data},
  author={Givens, Josh and Liu, Song and Reeve, Henry WJ},
  journal={arXiv preprint arXiv:2506.00557},
  year={2025}
}

@article{gottweis2025towards,
  title={Towards an AI co-scientist},
  author={Gottweis, Juraj and Weng, Wei-Hung and Daryin, Alexander and Tu, Tao and Palepu, Anil and Sirkovic, Petar and Myaskovsky, Artiom and Weissenberger, Felix and Rong, Keran and Tanno, Ryutaro and others},
  journal={arXiv preprint arXiv:2502.18864},
  year={2025}
}

@inproceedings{guo2025ideabench,
  title={Ideabench: Benchmarking large language models for research idea generation},
  author={Guo, Sikun and Shariatmadari, Amir Hassan and Xiong, Guangzhi and Huang, Albert and Kim, Myles and Williams, Corey M and Bekiranov, Stefan and Zhang, Aidong},
  booktitle={Proceedings of the 31st ACM SIGKDD Conference on Knowledge Discovery and Data Mining V. 2},
  pages={5888--5899},
  year={2025}
}

@inproceedings{hazra2025position,
  title={Position: AI Safety should prioritize the Future of Work},
  author={Hazra, Sanchaita and Majumder, Bodhisattwa Prasad and Chakrabarty, Tuhin},
  booktitle={Forty-second International Conference on Machine Learning Position Paper Track},
  year={2025}
}

@article{hua2025researchcodebench,
  title={Researchcodebench: Benchmarking llms on implementing novel machine learning research code},
  author={Hua, Tianyu and Hua, Harper and Xiang, Violet and Klieger, Benjamin and Truong, Sang T and Liang, Weixin and Sun, Fan-Yun and Haber, Nick},
  journal={arXiv preprint arXiv:2506.02314},
  year={2025}
}

@article{huang2025idea2plan,
  title={Idea2Plan: Exploring AI-Powered Research Planning},
  author={Huang, Jin and Cucerzan, Silviu and Jauhar, Sujay Kumar and White, Ryen W},
  journal={arXiv preprint arXiv:2510.24891},
  year={2025}
}

@article{huang2023mlagentbench,
  title={Mlagentbench: Evaluating language agents on machine learning experimentation},
  author={Huang, Qian and Vora, Jian and Liang, Percy and Leskovec, Jure},
  journal={arXiv preprint arXiv:2310.03302},
  year={2023}
}

@inproceedings{idahl2025openreviewer,
  title={Openreviewer: A specialized large language model for generating critical scientific paper reviews},
  author={Idahl, Maximilian and Ahmadi, Zahra},
  booktitle={Proceedings of the 2025 Conference of the Nations of the Americas Chapter of the Association for Computational Linguistics: Human Language Technologies (System Demonstrations)},
  pages={550--562},
  year={2025}
}

@misc{intology2025zochi,
  author       = {{Intology AI}},
  title        = {Zochi Technical Report},
  year         = {2025},
  howpublished = {\url{https://www.intology.ai/blog/zochi-tech-report}},
  note         = {Accessed: 2025-04-24}
}

@inproceedings{jansen2025codescientist,
  title={Codescientist: End-to-end semi-automated scientific discovery with code-based experimentation},
  author={Jansen, Peter and Tafjord, Oyvind and Radensky, Marissa and Siangliulue, Pao and Hope, Tom and Dalvi, Bhavana and Majumder, Bodhisattwa Prasad and Weld, Daniel S and Clark, Peter},
  booktitle={Findings of the Association for Computational Linguistics: ACL 2025},
  pages={13370--13467},
  year={2025}
}

@article{jiang2025aide,
  title={Aide: Ai-driven exploration in the space of code},
  author={Jiang, Zhengyao and Schmidt, Dominik and Srikanth, Dhruv and Xu, Dixing and Kaplan, Ian and Jacenko, Deniss and Wu, Yuxiang},
  journal={arXiv preprint arXiv:2502.13138},
  year={2025}
}

@article{kim2025position,
  title={Position: The AI conference peer review crisis demands author feedback and reviewer rewards},
  author={Kim, Jaeho and Lee, Yunseok and Lee, Seulki},
  journal={arXiv preprint arXiv:2505.04966},
  year={2025}
}

@article{kim2025train,
  title={Train for the worst, plan for the best: Understanding token ordering in masked diffusions},
  author={Kim, Jaeyeon and Shah, Kulin and Kontonis, Vasilis and Kakade, Sham and Chen, Sitan},
  journal={arXiv preprint arXiv:2502.06768},
  year={2025}
}

@article{lu2024ai,
  title={The ai scientist: Towards fully automated open-ended scientific discovery},
  author={Lu, Chris and Lu, Cong and Lange, Robert Tjarko and Foerster, Jakob and Clune, Jeff and Ha, David},
  journal={arXiv preprint arXiv:2408.06292},
  year={2024}
}

@article{nagarajan2025roll,
  title={Roll the dice \& look before you leap: Going beyond the creative limits of next-token prediction},
  author={Nagarajan, Vaishnavh and Wu, Chen Henry and Ding, Charles and Raghunathan, Aditi},
  journal={arXiv preprint arXiv:2504.15266},
  year={2025}
}

@article{nathani2025mlgym,
  title={Mlgym: A new framework and benchmark for advancing ai research agents},
  author={Nathani, Deepak and Madaan, Lovish and Roberts, Nicholas and Bashlykov, Nikolay and Menon, Ajay and Moens, Vincent and Budhiraja, Amar and Magka, Despoina and Vorotilov, Vladislav and Chaurasia, Gaurav and others},
  journal={arXiv preprint arXiv:2502.14499},
  year={2025}
}

@article{qiu2025ai,
  title={Ai idea bench 2025: Ai research idea generation benchmark},
  author={Qiu, Yansheng and Zhang, Haoquan and Xu, Zhaopan and Li, Ming and Song, Diping and Wang, Zheng and Zhang, Kaipeng},
  journal={arXiv preprint arXiv:2504.14191},
  year={2025}
}

@article{schmidgall2025agent,
  title={Agent laboratory: Using llm agents as research assistants},
  author={Schmidgall, Samuel and Su, Yusheng and Wang, Ze and Sun, Ximeng and Wu, Jialian and Yu, Xiaodong and Liu, Jiang and Moor, Michael and Liu, Zicheng and Barsoum, Emad},
  journal={Findings of the Association for Computational Linguistics: EMNLP 2025},
  pages={5977--6043},
  year={2025},
  publisher={Association for Computational Linguistics}
}

@article{siegel2024core,
  title={Core-bench: Fostering the credibility of published research through a computational reproducibility agent benchmark},
  author={Siegel, Zachary S and Kapoor, Sayash and Nagdir, Nitya and Stroebl, Benedikt and Narayanan, Arvind},
  journal={arXiv preprint arXiv:2409.11363},
  year={2024}
}

@article{snell2025conformal,
  title={Conformal prediction as bayesian quadrature},
  author={Snell, Jake C and Griffiths, Thomas L},
  journal={arXiv preprint arXiv:2502.13228},
  year={2025}
}

@article{starace2025paperbench,
  title={PaperBench: Evaluating AI's Ability to Replicate AI Research},
  author={Starace, Giulio and Jaffe, Oliver and Sherburn, Dane and Aung, James and Chan, Jun Shern and Maksin, Leon and Dias, Rachel and Mays, Evan and Kinsella, Benjamin and Thompson, Wyatt and others},
  journal={arXiv preprint arXiv:2504.01848},
  year={2025}
}

@article{tang2025ai,
  title={Ai-researcher: Autonomous scientific innovation},
  author={Tang, Jiabin and Xia, Lianghao and Li, Zhonghang and Huang, Chao},
  journal={arXiv preprint arXiv:2505.18705},
  year={2025}
}

@article{team2023gemini,
  title={Gemini: a family of highly capable multimodal models},
  author={Team, Gemini and Anil, Rohan and Borgeaud, Sebastian and Alayrac, Jean-Baptiste and Yu, Jiahui and Soricut, Radu and Schalkwyk, Johan and Dai, Andrew M and Hauth, Anja and Millican, Katie and others},
  journal={arXiv preprint arXiv:2312.11805},
  year={2023}
}

@article{weng2024cycleresearcher,
  title={Cycleresearcher: Improving automated research via automated review},
  author={Weng, Yixuan and Zhu, Minjun and Bao, Guangsheng and Zhang, Hongbo and Wang, Jindong and Zhang, Yue and Yang, Linyi},
  journal={arXiv preprint arXiv:2411.00816},
  year={2024}
}

@article{wu2025collabllm,
  title={Collabllm: From passive responders to active collaborators},
  author={Wu, Shirley and Galley, Michel and Peng, Baolin and Cheng, Hao and Li, Gavin and Dou, Yao and Cai, Weixin and Zou, James and Leskovec, Jure and Gao, Jianfeng},
  journal={arXiv preprint arXiv:2502.00640},
  year={2025}
}

@article{xiang2025scireplicate,
  title={Scireplicate-bench: Benchmarking llms in agent-driven algorithmic reproduction from research papers},
  author={Xiang, Yanzheng and Yan, Hanqi and Ouyang, Shuyin and Gui, Lin and He, Yulan},
  journal={arXiv preprint arXiv:2504.00255},
  year={2025}
}

@article{xin2026safereview,
  title={SafeReview: Defending LLM-based Review Systems Against Adversarial Hidden Prompts},
  author={Xin, Yuan and Weng, Yixuan and Zhu, Minjun and Ling, Ying and Qin, Chengwei and Hahn, Michael and Backes, Michael and Zhang, Yue and Yang, Linyi},
  journal={arXiv preprint arXiv:2604.26506},
  year={2026}
}

@article{yamada2025ai,
  title={The ai scientist-v2: Workshop-level automated scientific discovery via agentic tree search},
  author={Yamada, Yutaro and Lange, Robert Tjarko and Lu, Cong and Hu, Shengran and Lu, Chris and Foerster, Jakob and Clune, Jeff and Ha, David},
  journal={arXiv preprint arXiv:2504.08066},
  year={2025}
}

@article{rabby2025iterative,
  title={Iterative hypothesis generation for scientific discovery with Monte Carlo Nash equilibrium self-refining trees},
  author={Rabby, Gollam and Muhammed, Diyana and Mitra, Prasenjit and Auer, S{\"o}ren},
  journal={arXiv preprint arXiv:2503.19309},
  year={2025}
}

@inproceedings{yu2025tinyscientist,
  title={Tinyscientist: An interactive, extensible, and controllable framework for building research agents},
  author={Yu, Haofei and Xuan, Keyang and Li, Fenghai and Zhu, Kunlun and Lei, Zijie and Zhang, Jiaxun and Qi, Ziheng and Richardson, Kyle and You, Jiaxuan},
  booktitle={Proceedings of the 2025 Conference on Empirical Methods in Natural Language Processing: System Demonstrations},
  pages={558--590},
  year={2025}
}

@article{zhu2025ai,
  title={Ai scientists fail without strong implementation capability},
  author={Zhu, Minjun and Xie, Qiujie and Weng, Yixuan and Wu, Jian and Lin, Zhen and Yang, Linyi and Zhang, Yue},
  journal={arXiv preprint arXiv:2506.01372},
  year={2025}
}

@inproceedings{zhu2025deepreview,
  title={Deepreview: Improving llm-based paper review with human-like deep thinking process},
  author={Zhu, Minjun and Weng, Yixuan and Yang, Linyi and Zhang, Yue},
  booktitle={Proceedings of the 63rd Annual Meeting of the Association for Computational Linguistics (Volume 1: Long Papers)},
  pages={29330--29355},
  year={2025}
}

@article{wang2023scientific,
  title={Scientific discovery in the age of artificial intelligence},
  author={Wang, Hanchen and Fu, Tianfan and Du, Yuanqi and Gao, Wenhao and Huang, Kexin and Liu, Ziming and Chandak, Payal and Liu, Shengchao and Van Katwyk, Peter and Deac, Andreea and others},
  journal={Nature},
  volume={620},
  number={7972},
  pages={47--60},
  year={2023},
  publisher={Nature Publishing Group UK London}
}

@article{bran2023chemcrow,
  title={Chemcrow: Augmenting large-language models with chemistry tools},
  author={Bran, Andres M and Cox, Sam and Schilter, Oliver and Baldassari, Carlo and White, Andrew D and Schwaller, Philippe},
  journal={arXiv preprint arXiv:2304.05376},
  year={2023}
}

@article{boiko2023autonomous,
  title={Autonomous chemical research with large language models},
  author={Boiko, Daniil A and MacKnight, Robert and Kline, Ben and Gomes, Gabe},
  journal={Nature},
  volume={624},
  number={7992},
  pages={570--578},
  year={2023},
  publisher={Nature Publishing Group UK London}
}

@article{keya2026sci,
  title={Sci-idea: Context-aware scientific ideation using token and sentence embeddings},
  author={Keya, Farhana and Rabby, Gollam and Auer, S{\"o}ren and Vahdati, Sahar and Mitra, Prasenjit and Jaradeh, Mohamad Yaser},
  journal={Machine Learning},
  volume={115},
  number={5},
  pages={103},
  year={2026},
  publisher={Springer}
}

@article{li2024mlr,
  title={Mlr-copilot: Autonomous machine learning research based on large language models agents},
  author={Li, Ruochen and Patel, Teerth and Wang, Qingyun and Du, Xinya},
  journal={arXiv preprint arXiv:2408.14033},
  year={2024}
}

@article{seo2025paper2code,
  title={Paper2code: Automating code generation from scientific papers in machine learning},
  author={Seo, Minju and Baek, Jinheon and Lee, Seongyun and Hwang, Sung Ju},
  journal={arXiv preprint arXiv:2504.17192},
  year={2025}
}

@article{ghafarollahi2025sciagents,
  title={SciAgents: automating scientific discovery through bioinspired multi-agent intelligent graph reasoning},
  author={Ghafarollahi, Alireza and Buehler, Markus J},
  journal={Advanced Materials},
  volume={37},
  number={22},
  pages={2413523},
  year={2025},
  publisher={Wiley Online Library}
}

@article{kon2025curie,
  title={Curie: Toward rigorous and automated scientific experimentation with ai agents},
  author={Kon, Patrick Tser Jern and Liu, Jiachen and Ding, Qiuyi and Qiu, Yiming and Yang, Zhenning and Huang, Yibo and Srinivasa, Jayanth and Lee, Myungjin and Chowdhury, Mosharaf and Chen, Ang},
  journal={arXiv preprint arXiv:2502.16069},
  year={2025}
}
\newpage
% \section{NeurIPS Evaluation Form}
\label{appx:neurips_eval_form}

% \input{checklist}

% \newpage
\appendix

\section{APPENDICES}

\subsection{Features of MLReplicate}
\label{sec:Features_of_MLReplicate}

\paragraph{High-quality and Trustworthy.}
MLReplicate is built from a curated set of outstanding ICML 2025 papers, ensuring all tasks are grounded in high-quality, peer-reviewed research. Each paper is converted into a standardized ground-truth specification that preserves core elements such as hypotheses, experimental design, and evaluation criteria, reducing noise and ambiguity for consistent assessment. The benchmark also combines automated and expert human evaluation, enhancing robustness while avoiding over-reliance on LLM-based judgments.

\paragraph{Challenging and Novel.}
Rather than simplified proxy tasks, MLReplicate reflects the complexity of real-world
scientific workflows, requiring systems to perform end-to-end research generation problem
formulation, experimentation, and manuscript writing under conditions of partial
specification, introducing inherent ambiguity and non-trivial design decisions. By covering
diverse machine learning subfields and research types, MLReplicate presents a challenging and previously underexplored evaluation setting
for autonomous research systems.
 
\paragraph{Flexible and Extensible.}
MLReplicate adopts a modular design that supports straightforward integration of new
systems, datasets, and evaluation protocols. The standardized groundtruth representation
is adaptable to multiple input formats, ensuring compatibility with a wide range of agents
architectures: pipeline-based, multi-agent, and tree-search, and the benchmark can be
extended to domains beyond machine learning, with support for additional evaluation
dimensions such as experimental verification, robustness checks, or domain-specific
metrics.
 
\paragraph{Lightweight and Efficient.}
Despite its end-to-end scope, MLReplicate is intentionally lightweight and computationally
manageable: a small but carefully curated dataset enables rapid iteration and reproducibility
experimentation without large-scale compute, while systematic tracking of runtime, cost,
and resource usage provides a practical perspective on system efficiency, striking a balance between realism and accessibility suitable for both research exploration and
standardized evaluation.

\subsection{Large Language Models (LLMs)}
\label{sec:llm_configs}

\begin{table}[h]
\centering
\scriptsize
\caption{\textbf{Configurable language models supported by each evaluated autonomous research system.}}
\label{tab:llm_configs}
\renewcommand{\arraystretch}{1.15}
\setlength{\tabcolsep}{4pt}
\begin{tabularx}{\columnwidth}{p{3cm} X}
\toprule
\textbf{System} & \textbf{Models}  \\
\midrule
AI Scientist v1
& \texttt{gpt-4o-2024-05-13}, \texttt{claude-3-5-sonnet-20241022} \\

AI Scientist v2
& \texttt{gpt-4o-2024-05-13}, \texttt{o1-preview-2024-09-12}, \texttt{gpt-4o-2024-11-20}, \texttt{o3-mini-2025-01-31}  \\

Agent Laboratory
& \texttt{o1-preview}, \texttt{o1-mini}, \texttt{gpt-4o}  \\

AI Researcher
& \texttt{gpt-4o}, \texttt{gpt-4o-mini-2024-07-18}, \texttt{o1-mini}, \texttt{claude-3-5-sonnet-20241022}, \texttt{claude-3-5-haiku-20241022}, \texttt{deepseek-chat}, \texttt{deepseek-reasoner}  \\

Tiny Scientist
& \texttt{gpt-4o} \texttt{gpt-4o-mini}, \texttt{o1-mini}, \texttt{o1-preview},
  \texttt{gpt-3.5-turbo}, \texttt{claude-3-haiku}, \texttt{claude-3-sonnet},
  \texttt{claude-3.5-sonnet}, \texttt{deepseek-chat}, \texttt{deepseek-reasoner},
  \texttt{gemini-1.5-flash}, \texttt{gemini-1.5-pro}, \texttt{llama-3.1-40b}  \\

CycleResearcher
& \texttt{CycleResearcher-12B}, \texttt{CycleResearcher-72B}, \texttt{CycleResearcher-123B}  \\
\bottomrule
\end{tabularx}
\end{table}

% % \begin{figure}[t]
% %     \centering
% %     \includegraphics[width=0.7\linewidth]{img/05_generated_papers_pie_chart.pdf}
% %     \caption{Pie Chart representing the percentage of paper generated, failed and desk rejected}
% %     \label{fig:pie_chart_analysis}
   
% % \end{figure}

% \begin{wrapfigure}{r}{0.4\columnwidth}
%     \centering
%     % \vspace{-10pt}
%     \includegraphics[width=0.38\columnwidth]{img/05_generated_papers_pie_chart.pdf}
%     % \vspace{-10pt}
%     \caption{Pie Chart representing the percentage of paper generated, failed and desk rejected}
%     \label{fig:pie_chart_analysis}
%     \vspace{-1cm}
% \end{wrapfigure}

% \textcolor{green}{
% Figure \ref{fig:pie_chart_analysis} illustrates the distribution of papers across failure, desk rejection, and successful evaluation stages.
% A total of 48 papers were initially generated from all six autonomous research systems. Among these, 8 were desk-rejected for not meeting the minimum two-page requirement, all of which were produced by the AI Scientist V1 system. In addition, three experimental runs failed to generate any output from their respective autonomous research systems. After applying these filters, 37 papers remained eligible for further consideration. From this subset, 10 papers were accepted by the AI ICAIS conference and proceeded to human evaluation. }

% \newpage

\begin{table}[h]
\centering
\caption{\textbf{Representative human reviewer comments per autonomous research systems.} Feedback is condensed from free-text evaluations focusing on hallucinations, limitations, and ethical concerns.}
\label{tab:qualitative}
\renewcommand{\arraystretch}{1.15}
\setlength{\tabcolsep}{6pt}
\small
\begin{tabular}{p{3cm} p{10cm}}
\toprule
\textbf{System} & \textbf{Reviewer Comments} \\
\midrule

\textsc{Agent Laboratory}
& ``Evaluation conducted exclusively on synthetic datasets.''; 
``No real-world validation.''; 
``Critical reproducibility details are missing.'' \\

\textsc{AI Researcher}
& ``Experimental section contains placeholders (e.g., [Dataset Name], [Metric1], [Metric2]) instead of concrete values.''; 
``Fabricated experimental results; non-existent ablation studies.'' \\

\textsc{CycleResearcher}
& ``Contradictory speedup claims.''; 
``Broken references to non-existent content.''; 
``Key implementation details are missing.''; 
``Duplicate tables detected.'' \\

\textsc{Tiny Scientist}
& ``Dataset--task mismatch and complete model failure.''; 
``Maximizes a fabricated metric that is not well-defined.'' \\

\bottomrule
\end{tabular}
\end{table}
\subsection{Qualitative Analysis}
\label{sec:human_reviewer_comments}

Fabricated or unsupported experimental content emerged as a consistent pattern across all four systems that reached human evaluation. \textsc{Agent Laboratory} submissions were most frequently criticized for conducting evaluations exclusively on synthetic datasets and for making broad generalization claims without adequate empirical support. \textsc{AI Researcher} attracted the most severe concerns: reviewers repeatedly identified unresolved template placeholders (e.g., \texttt{[Dataset Name]}, \texttt{[Metric 1]}, \texttt{[Metric 2]}) and references to a causal-inference module that was mentioned in the text but entirely absent from the implementation. \textsc{Tiny Scientist} was flagged for systematic dataset-task mismatches and for employing an ill-defined metric as a proxy for scientific novelty. \textsc{CycleResearcher}, while producing more fluent text, suffered from notable reproducibility issues, including references to non-existent content and contradictory quantitative claims within the same manuscript. \textsc{AI Scientist-v2}, which did not pass the automated screening stage, showed early indications of difficulty in generating coherent long-form experimental reports without heavy human intervention. Collectively, these qualitative observations strongly reinforce the quantitative findings. Although the generated manuscripts often appear structurally plausible and professionally formatted, they remain scientifically unreliable. Hallucinations, methodological inconsistencies, and missing implementation details are pervasive across all systems, regardless of their architectural design or computational budget. Representative reviewer comments highlighting these issues are summarized in \autoref{tab:qualitative}.

\begin{table*}[h]
\centering
\small
\setlength{\tabcolsep}{4pt}
\caption{\textbf{Autonomous research systems considered for \textsc{MLReplicate}.} Systems above the line are evaluated; those below are excluded with rationale.}
\label{tab:systems}
\begin{tabular}{p{3cm} p{3.6cm} p{3.2cm} c}
\toprule
\textbf{System} & \textbf{Architecture} & \textbf{Task Scope} & \textbf{Open Source} \\
\midrule
\textsc{AI Scientist-v2}   & Agentic tree search                      & Full pipeline       & Yes \\
\textsc{AI Researcher}     & Multi-agent                              & Full pipeline       & Yes \\
\textsc{Agent Laboratory}  & Multi-agent                              & Full pipeline       & Yes \\
\textsc{CycleResearcher}   & fine-tuned LLMs         & Writing \& review   & Yes \\
\textsc{Tiny Scientist}    & Sequential workflow                  & Full pipeline       & Yes \\
\textsc{AI Scientist v1}   & Sequential workflow                                 & Full pipeline       & Yes \\

\midrule
\textsc{Zochi}             & -                     & Full pipeline       & No  \\
\textsc{AI Co-scientist}   & Multi-agent                              & Ideation \& planning & No  \\
%\textsc{Dolphin}           & Closed-loop pipeline                     & Full pipeline       & Yes \\
\textsc{AIDE}              & Agentic tree search                      & ML engineering      & Yes \\
\textsc{CodeScientist}     & Semi-automated pipeline                  & code experimentation & Yes \\
\textsc{Curie}             & Multi-agent                              & Experimentation     & Yes \\
\textsc{Paper2Code}        & Multi-agent pipeline                     & Code from paper     & Yes \\
\textsc{MLR-Copilot}       & Sequential workflow                                & Full pipeline       & Yes \\
\textsc{SciAgents}         & Multi-agent (graph-based)                & Hypothesis generation & Yes \\
\bottomrule
\end{tabular}
\end{table*}

\subsection{Autonomous Research Systems}

Table~\ref{tab:systems} summarizes all autonomous research systems considered for
\textsc{MLReplicate}, along with the rationale for exclusion where applicable. \textsc{Zochi} \citep{intology2025zochi} and \textsc{AI Co-Scientist} \citep{gottweis2025towards} are excluded as both are closed-source and therefore cannot be reliably deployed or reproduced in a standardized evaluation setting. \textsc{AI Co-Scientist} is further limited in scope, addressing only
ideation and planning rather than the full research pipeline. \textsc{AIDE} \citep{jiang2025aide} and \textsc{CodeScientist} \citep{jansen2025codescientist} are excluded due to their narrow task scope, as they are restricted to code experimentation. Similarly, \textsc{Curie} \citep{kon2025curie} focuses solely on the experimentation phase, and \textsc{Paper2Code} \citep{seo2025paper2code} is limited to code generation from existing papers; neither constitutes a complete autonomous research pipeline. \textsc{SciAgents} \citep{ghafarollahi2025sciagents} is likewise excluded, as it addresses only hypothesis generation. Finally, despite nominally covering idea generation, experiment implementation, and execution, \textsc{MLR-Copilot} \citep{li2024mlr} is excluded as it requires heavy human steering and does not meet the autonomy threshold throughout the experimentation phase.

\subsection{AI Conference}
\label{sec:ai_conference}
The automated review results in our experiments were obtained using the platform developed for the \textit{1st International Conference on AI Scientists} (ICAIS 2025)\footnote{ICAIS 2025 link: \url{https://airaxiv.com/events/icais2025/?decision=spotlight_accept}}, organized by the Zhongguancun Academy, the Zhongguancun Institute of Artificial Intelligence, Tsinghua University, Westlake University, and the University of Chicago. We leveraged the ICAIS infrastructure because it enables the generation of multiple automated reviews at scale while maintaining practical usability. Looking ahead, we believe this setup can be readily adapted into a fully automated review conference powered by more advanced open-source review models for more transparency of the benchmark evaluation.

\newpage

% \section{Benchmark Details}

\section{System configurations}
\label{sec:Benchmark_Metadata_Format}

\subsection{\textsc{AI Scientist-v1}}
\paragraph{Credentials \& Hardware.} 
Uses one or more NVIDIA A100 GPUs (40–80GB GPU memory per GPU depending on availability), 16 CPU cores, and 64GB of system memory, with optional multi-GPU scaling for parallel experiment generation and evaluation.

\begin{tcolorbox}[
    title=Input,
    colback=blue!5!white,
    colframe=blue!75!black,
    enhanced,
    breakable,
]
\begin{Verbatim}[breaklines=true, breakanywhere=true]
{
    "experiment_name": "[Experiment Name]",
    "template": "[Template Name (e.g., 2d_diffusion, nanogpt, grokking)]",
    "title": "[Proposed Paper Title]",
    "keywords": ["keyword1", "keyword2", "keyword3"],
    "tldr": "[One-sentence summary of the research idea]",
    "abstract": "[Full research abstract describing the idea, method, and expected contribution]",
    "num_ideas": 1,
    "model": "[LLM used for generation, e.g., gpt-4o, claude-3.5]"
}
\end{Verbatim}
\end{tcolorbox}

% \begin{tcolorbox}[
%     title=Output,
%     colback=blue!5!white,
%     colframe=blue!75!black,
%     enhanced,
%     breakable,
% ]
% \begin{Verbatim}[breaklines=true, breakanywhere=true]
% {
%     "experiment_name": "[Experiment Name]",

%     "generated_paper": {
%         "title": "[Final Paper Title]",
%         "latex_source": "paper.tex",
%         "pdf": "paper.pdf",
%         "sections": [
%             "abstract",
%             "introduction",
%             "method",
%             "experiments",
%             "results",
%             "discussion",
%             "conclusion"
%         ],
%         "figures": [
%             "figures/figure1.png",
%             "figures/figure2.png"
%         ],
%         "bibliography": "references.bib"
%     },

%     "experimental_results": {
%         "metrics": {
%             "accuracy": "[value]",
%             "loss": "[value]",
%             "other_metrics": "[values]"
%         },
%         "logs": "logs/training.log",
%         "plots": [
%             "results/loss_curve.png",
%             "results/accuracy_curve.png"
%         ],
%         "checkpoints": [
%             "models/checkpoint_last.pt"
%         ]
%     },

%     "review": {
%         "summary": "[AI-generated peer review summary]",
%         "strengths": ["strength1", "strength2"],
%         "weaknesses": ["weakness1", "weakness2"],
%         "score": "[0-10]"
%     }
% }
% \end{Verbatim}
% \end{tcolorbox}

\subsection{\textsc{AI Scientist-v2}}
\paragraph{Credentials \& Hardware.} Uses two NVIDIA L40S GPUs (48GB GPU memory each), 8 CPU cores, and 12GB of system memory.

\begin{tcolorbox}[
    title=Input,
    colback=blue!5!white,
    colframe=blue!75!black,
    enhanced,
    breakable,
]
\begin{Verbatim}[breaklines=true, breakanywhere=true]
{
    "title": [Title],
    "keywords": [Keywords],
    "tldr": [TL;DR],
    "abstract": [Abstract]
}
\end{Verbatim}
\end{tcolorbox}

\begin{tcolorbox}[
    title=Configuration YAML File,
    colback=blue!5!white,
    colframe=blue!75!black,
    enhanced,
    breakable,
]
\begin{Verbatim}[breaklines=true, breakanywhere=true]
{
    "data_dir": [Data directory],
    "preprocess_data": [True/False],

    "goal": [Goal],
    "eval": [Evaluation settings],

    "log_dir": [Log directory],
    "workspace_dir": [Workspace directory],

    "copy_data": [True/False],

    "exp_name": [Experiment name],

    "exec": {
        "timeout": [Timeout],
        "agent_file_name": [File name],
        "format_tb_ipython": [True/False]
    },

    "generate_report": [True/False],

    "report": {
        "model": [Model],
        "temp": [Temperature]
    },

    "experiment": {
        "num_syn_datasets": [Number of synthetic datasets]
    },

    "debug": {
        "stage4": [True/False]
    },

    "agent": {
        "type": [Agent type],
        "num_workers": [Number of workers],

        "stages": {
            "stage1_max_iters": [Max iterations],
            "stage2_max_iters": [Max iterations],
            "stage3_max_iters": [Max iterations],
            "stage4_max_iters": [Max iterations]
        },

        "steps": [Number of improvement steps],
        "k_fold_validation": [True/False],
        
        "multi_seed_eval": {
            "num_seeds": [Number of seeds]
        },

        "expose_prediction": [True/False],
        "data_preview": [True/False],

        "code": {
            "model": [Model],
            "temp": [Temperature],
            "max_tokens": [Max tokens]
        },

        "feedback": {
            "model": [Model],
            "temp": [Temperature],
            "max_tokens": [Max tokens]
        },

        "vlm_feedback": {
            "model": [Model],
            "temp": [Temperature],
            "max_tokens": [Max tokens]
        },

        "search": {
            "max_debug_depth": [Max depth],
            "debug_prob": [Probability],
            "num_drafts": [Number of drafts]
        }
    }
}
\end{Verbatim}
\end{tcolorbox}

% \begin{tcolorbox}[
%     title=Output,
%     colback=blue!5!white,
%     colframe=blue!75!black,
%     enhanced,
%     breakable,
% ]
% \begin{Verbatim}[breaklines=true, breakanywhere=true]
% {
%     "output_directory_structure": {
%         "figures/": "Generated plots, diagrams, and visualizations",
%         "latex/": "LaTeX source files for paper generation",
%         "logs/": "Experimental logs and results",
%         "experimental_codes/": "All scripts and implementation code",
        
%         "configuration.yaml": "Experiment configuration and hyperparameters",
%         "cached_citations.bib": "Cached bibliography entries",

%         "idea.json": {
%             "Name": "",
%             "Title": "",
%             "Short Hypothesis": "",
%             "Related Work": "",
%             "Experiments": "",
%             "Risk Factors": "",
%             "Limitations": ""
%         },

%         "token_tracker_interactions.json": "Record of all prompts and model interactions",
%         "token_tracker.json": "Token usage statistics per model and total cost tracking"
%     }
% }
% \end{Verbatim}
% \end{tcolorbox}

\subsection{\textsc{Agent Laboratory}}

\paragraph{Credentials \& Hardware.} An example YAML file with \texttt{task-notes}, \texttt{data-preparation}, \texttt{running-experiments}, \texttt{results-interpretation}, and \texttt{report-writing}, generated by \texttt{gpt-5} by inputting the corresponding paper from the dataset. It requires \texttt{OPENAI\_API\_KEY} or \texttt{DEEPSEEK\_API\_KEY}. We defaulted to model \texttt{o3\_mini} through out the \textsc{Agent Laboratory} pipeline.

\begin{tcolorbox}[
        title=Input YAML File,
        colback=blue!5!white,
        colframe=blue!75!black,
        enhanced,    % Required for advanced skin features
        breakable,   % This allows the box to split across pages
        ]
\begin{Verbatim}[breaklines=true, breakanywhere=true]
copilot-mode: True
research-topic: "..."

api-key: "YOUR_OPENAI_API_KEY_HERE"
llm-backend: "o3-mini"
# Literature review backend
lit-review-backend: "o3-mini"

# Base language
language: "English"

# Number of arxiv papers to lit review
num-papers-lit-review: 5
# Total number of papers to write in sequence
num-papers-to-write: 1
# Do you want to run multiple agent labs in parallel?
parallel-labs: False

# Total mle-solver steps per lab
mlesolver-max-steps: 2
# Total paper-solver steps per lab
papersolver-max-steps: 2
# The lab index for this lab (used for parallel runs)
lab-index: 1
# If you want to load an existing save
load-existing: False
# If fail, run exception?
except-if-fail: False
# Compile latex into PDFs during paper-solver
compile-latex: False

task-notes:
  plan-formulation:
    - '...'
    - '...'
    - '...'

  data-preparation:
    - '...'
    - '...'
    - '...'

  running-experiments:
    - '...'
    - '...'
    - '...'

  results-interpretation:
    - '...'
    - '...'
    - '...'

  report-writing:
    - '...'
    - '...'
    - '...'

\end{Verbatim}

\end{tcolorbox}

\subsection{\textsc{AI Researcher}}

\paragraph{Credentials \& Hardware.}
Requires \texttt{OPENAI\_API\_KEY} (mandatory). \texttt{GITHUB\_AI\_TOKEN} is optional and was used for repository search by the Knowledge Acquisition Agent. Requires a Docker or Singularity sandbox for the Code Agent. Each run was scheduled on a SLURM node with 8 CPU cores, 32\,GB RAM, and one NVIDIA RTX\,3090.

\begin{tcolorbox}[colback=blue!5!white, colframe=blue!75!black, boxrule=0.4pt, left=6pt, right=6pt, top=4pt, bottom=4pt, title={\small\textbf{Input}}, fonttitle=\small]
One JSON file per run under Level\,2 (Reference-Based Ideation), containing the target paper title and a ranked list of source papers with their references, types, justifications, and usage descriptions. Each source paper entry also carries metadata including authors, year and abstract Example (abbreviated):
\vspace{4pt}

\begin{Verbatim}[breaklines=true, breakanywhere=true, fontsize=\small]
{
  "target": "Conformal Prediction as Bayesian Quadrature",
  "source_papers": [
    {
      "reference": "Vovk, V., Gammerman, A., and Shafer, G. 
                    Algorithmic Learning in a Random World.",
      "rank": 1,
      "type": ["methodological"],
      "justification": "Seminal work introducing conformal 
                        prediction methods.",
      "usage": "Extended to develop the proposed framework."
    },
    ...
  ],
  "authors": ["Jake C. Snell", "Thomas L. Griffiths"],
  "year": 2025,
  "abstract": "..."
}
\end{Verbatim}

The benchmark instances were constructed using a multi-step pipeline: papers were crawled from arXiv, their PDFs retrieved, and an LLM-based innovation graph was generated to identify and rank source papers by methodological relevance. The pipeline for benchmark construction is available in repository
\end{tcolorbox}

% \begin{tcolorbox}[colback=blue!5!white, colframe=blue!75!black, boxrule=0.4pt, left=6pt, right=6pt, top=4pt, bottom=4pt, title={\small\textbf{Output}}, fonttitle=\small]
% One directory per run containing: per-agent state checkpoints, a retrieved-paper memory store, a generated codebase, terminal logs, and a \texttt{paper/} folder with the final PDF and intermediate drafting traces.
% \end{tcolorbox}

\begin{tcolorbox}[colback=blue!5!white, colframe=blue!75!black, boxrule=0.4pt, left=6pt, right=6pt, top=4pt, bottom=4pt, title={\small\textbf{Key Configuration}}, fonttitle=\small]
\begin{tabular}{@{}ll@{}}
Primary model      & \texttt{gpt-4o-2024-08-06} \\
Lightweight calls  & \texttt{gpt-4o-mini-2024-07-18} \\
Embeddings         & \texttt{text-embedding-3-small} \\
Function calling   & enabled \\
Operating mode     & Level\,2 (Reference-Based Ideation) \\
\end{tabular}
\end{tcolorbox}

\subsection{\textsc{CycleResearcher}}

\paragraph{Credentials \& Hardware.} Requires at least 23GB of GPU memory with input text and relevant bbl files.

\begin{tcolorbox}[
        title=Input,
        colback=blue!5!white,
        colframe=blue!75!black,
        enhanced,    % Required for advanced skin features
        breakable,   % This allows the box to split across pages
        ]
\begin{Verbatim}[breaklines=true, breakanywhere=true]
{
    "title": "Conformal Prediction as Bayesian Quadrature",
    "references": "dataset/latex_folders/11432_Conformal_Prediction_as_/main.bbl"
}
\end{Verbatim}
\end{tcolorbox}

with the \texttt{main.bbl} file containing the following relevant references from the paper.

\begin{tcolorbox}[
        title=bbl file contents,
        colback=blue!5!white,
        colframe=blue!75!black,
        enhanced,    % Required for advanced skin features
        breakable,   % This allows the box to split across pages
        ]
\begin{Verbatim}[breaklines=true, breakanywhere=true]
\begin{thebibliography}{29}
\providecommand{\natexlab}[1]{#1}
\providecommand{\url}[1]{\texttt{#1}}
\expandafter\ifx\csname urlstyle\endcsname\relax
  \providecommand{\doi}[1]{doi: #1}\else
  \providecommand{\doi}{doi: \begingroup \urlstyle{rm}\Url}\fi

\bibitem[Aitchison \& Dunsmore(1975)Aitchison and
  Dunsmore]{aitchison1975statistical}
Aitchison, J. and Dunsmore, I.~R.
\newblock \emph{Statistical Prediction Analysis}.
\newblock New York: Cambridge University Press, 1975.
....

\end{thebibliography}
\end{Verbatim}
\end{tcolorbox}

\subsection{\textsc{Tiny Scientist}}

\paragraph{Credentials \& Hardware.}
Requires \texttt{OPENAI\_API\_KEY} (mandatory), \texttt{GITHUB\_TOKEN} (optional), and Docker for sandboxed experiment execution (falls back to local execution if unavailable). 

\begin{tcolorbox}[colback=blue!5!white, colframe=blue!75!black, boxrule=0.4pt, left=6pt, right=6pt, top=4pt, bottom=4pt, title={\small\textbf{Input}}, fonttitle=\small]
One research intent string per run, following the format: \textit{``The main research goal is to [GOAL]. The main research question is: [QUESTION].''}
The 8 intents were generated from published abstracts via a zero-shot \texttt{gpt-4o} prompt. Example:
\vspace{4pt}

\textit{``The main research goal is to reform the peer review process in AI conferences to enhance review quality and reviewer accountability. The main research question is: How can implementing a bi-directional feedback loop and a structured reviewer reward system improve reviewer accountability and foster a sustainable peer review ecosystem?''}
\end{tcolorbox}

% \begin{tcolorbox}[colback=gray!5, colframe=gray!40, boxrule=0.4pt, left=6pt, right=6pt, top=4pt, bottom=4pt, title={\small\textbf{Output}}, fonttitle=\small]
% One directory per run containing: run metadata, Thinker output, per-stage timing and cost logs, a human-readable summary, generated experiment code, experiment results, and a \texttt{latex/} folder with the final manuscript and retrieved citations.
% \end{tcolorbox}

\begin{tcolorbox}[colback=blue!5!white, colframe=blue!75!black, boxrule=0.4pt, left=6pt, right=6pt, top=4pt, bottom=4pt, title={\small\textbf{Key Configuration}}, fonttitle=\small]
\begin{tabular}{@{}ll@{}}
Model            & \texttt{gpt-4o} \\
Temperature      & 0.75 \\
Budget cap       & \$10.00 USD per run \\

\end{tabular}
\end{tcolorbox}

\newpage

\section{Human Evaluation}

\begin{figure}[htbp]
    \centering
    
    % \begin{subfigure}{0.48\linewidth}
    %     \includegraphics[width=\linewidth]{img/human_eval/human_eval_form_7.png}
    % \end{subfigure}
    \begin{subfigure}{0.48\linewidth}
        \includegraphics[width=\linewidth]{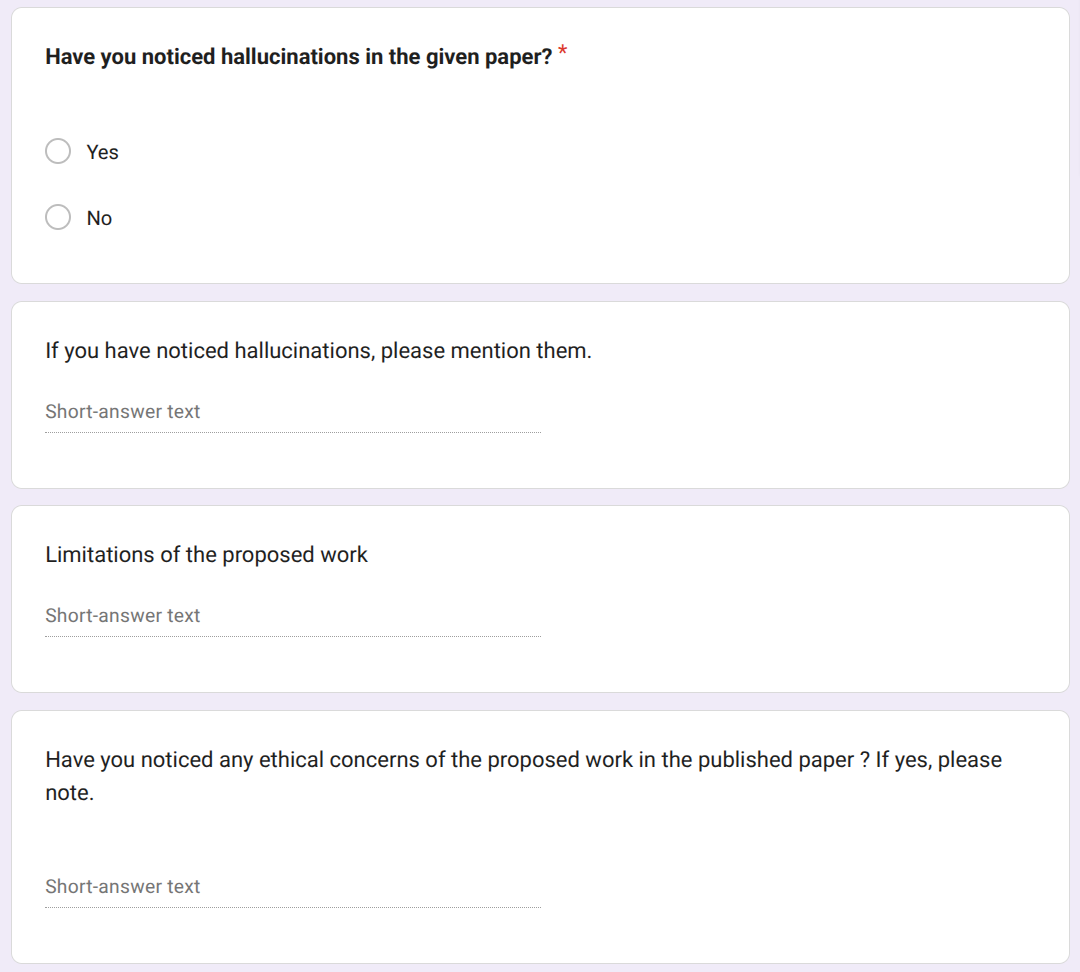}
    \end{subfigure}    
    \begin{subfigure}{0.48\linewidth}
        \includegraphics[width=\linewidth]{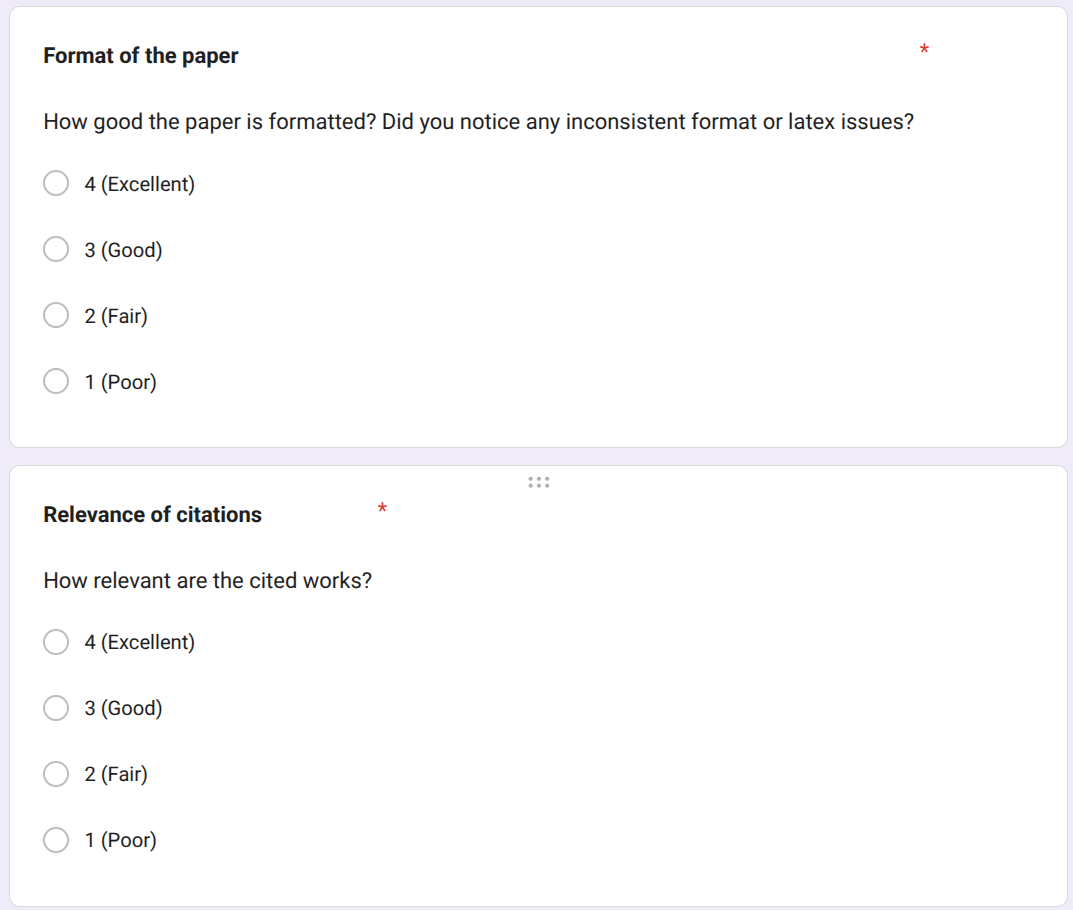}
    \end{subfigure}

    \vspace{0.5em}
    \begin{subfigure}{0.48\linewidth}
        \includegraphics[width=\linewidth]{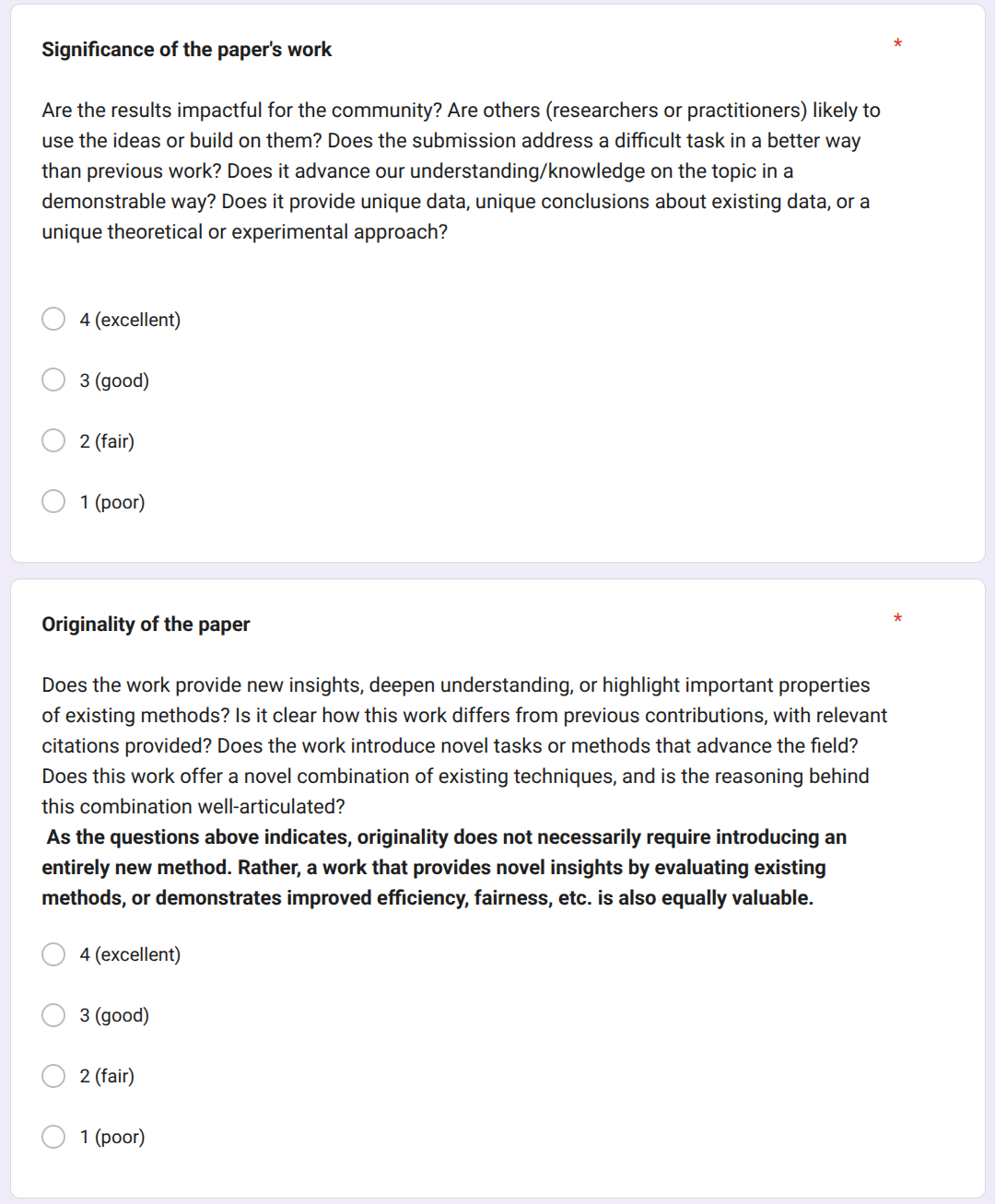}
    \end{subfigure}    
    \begin{subfigure}{0.48\linewidth}
        \includegraphics[width=\linewidth]{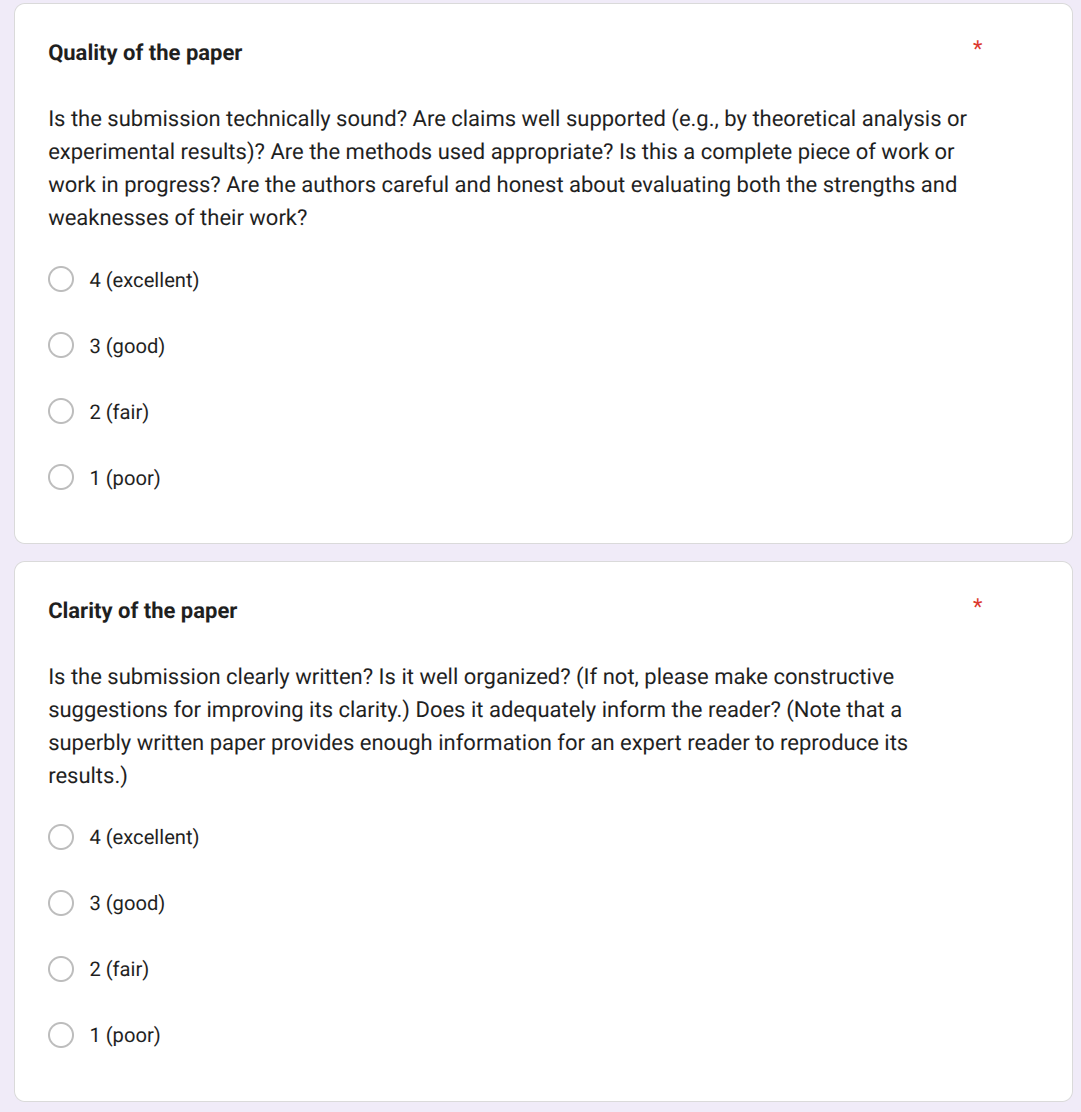}
    \end{subfigure}

    % \vspace{0.5em}
    
    % \begin{subfigure}{0.48\linewidth}
    %     \includegraphics[width=\linewidth]{img/human_eval/human_eval_form_1.png}
    % \end{subfigure}

    \caption{Excerpts from the human evaluation forms, designed in the style of NeurIPS review templates, illustrating the structured assessment framework used by expert reviewers.}
    \label{fig:human_eval_all}
\end{figure}

\autoref{fig:human_eval_all} showcases excerpts of the human evaluation form. It is designed to evaluate according to the NeurIPS format. In addition, we also request to provide the hallucinations and ethical considerations of the autonomous research system-generated papers.

% \section{Prompts}

\newpage

\section{Error Breakdown}
\label{sec:erroranalysis}

\begin{table}[h]
\centering
\caption{\textbf{Technical Taxonomy of Multi-Agent Pipeline Failures.} We categorize the observed errors during the execution of the Tiny Scientist framework, mapping high-level agentic failures to specific log symptoms and their downstream impact on the research pipeline.}
\label{tab:error_analysis}
\small
\begin{tabularx}{\textwidth}{@{}l p{3cm} p{4cm} X @{}}
\toprule
\textbf{Category} & \textbf{Root Cause} & \textbf{Log Symptoms / Error Codes} & \textbf{Pipeline Impact} \\ \midrule
\textbf{Cross-Modal Hallucination} & Incongruence between Thinker plan and dataset modality. & \texttt{ValueError: Input 0 of layer "cnn" is incompatible...} & \textbf{Fatal:} Immediate crash during model initialization. \\ \addlinespace
\textbf{Hardware Factory Conflict} & Redundant initialization of GPU backends across iterations. & \texttt{AbortedError: Unable to register cuFFT/cuDNN/cuBLAS factory.} & \textbf{Instability:} Potential memory leaks and non-deterministic results. \\ \addlinespace
\textbf{Logic Implementation Drift} & Semantic planning errors inherited from the Thinker stage. & \texttt{AttributeError: 'NoneType' object has no attribute 'shape'.} & \textbf{High Latency:} Forces Coder into futile 4-iteration debug loops. \\ \addlinespace
\textbf{Data-Model Contract Breach} & Mismatch in tensor shapes during the training phase. & \texttt{InvalidArgumentError: Matrix size-incompatible: [batch, text] vs [image\_dims].} & \textbf{Execution Halt:} Training terminates at the first epoch. \\ \addlinespace
\textbf{System Boundary Fault} & Attempt to access non-provisioned local assets or weights. & \texttt{FileNotFoundError} or \texttt{ModuleNotFoundError} (external assets). & \textbf{Fatal:} Environment setup failure prevents execution. \\ \addlinespace
\textbf{Constraint Exhaustion} & Reaching the ceiling of the agentic retry budget. & \texttt{SystemExit: Maximum retries (2/2) exceeded for stage 'Coder'.} & \textbf{Abortion:} Experiment marked as Failed despite valid syntax. \\ \addlinespace
\textbf{API Parameter Mismatch} & Usage of deprecated or incorrect Keras/TF arguments. & \texttt{TypeError: \_\_init\_\_() got an unexpected keyword argument.} & \textbf{Recoverable:} Corrected via internal Coder self-correction. \\ \bottomrule
\end{tabularx}
\end{table}

To understand the failure modes of autonomous research systems at the execution level, we analyze runtime logs from the Tiny Scientist pipeline and identify a taxonomy of seven recurring technical failure categories, summarized in \autoref{tab:error_analysis}. This taxonomy illuminates the fundamental tension between probabilistic planning and
deterministic execution in multi-agent research systems.
 
\paragraph{Taxonomy of Failures.}
Cross-modal hallucination is the most consequential failure mode: the Thinker stage proposes architectures incompatible with the input data modality, for instance, planning a CNN-based image feature extractor for the text-only \texttt{yelp\_polarity} dataset, causing an immediate fatal crash during model initialization.
Logical implementation drift arises when code is syntactically correct but semantically misaligned with the data contract: the Coder initializes a multimodal model expecting three input streams (text, image, audio) while the data generator yields a single text tensor, forcing futile debug iterations.
Hardware factory conflicts occur when the GPU backend (cuFFT, cuDNN, cuBLAS) is redundantly re-initialized across agent iterations, introducing memory instability and non-deterministic results.
Data-model contract violations produce hard crashes during \texttt{model.fit()} due to tensor shape mismatches, for example, when the pipeline expects multidimensional audio spectrograms, absent from an NLP benchmark.
System boundary faults\ arise when agents attempt to load local assets or pre-trained weights not provisioned in the sandbox, resulting in \texttt{FileNotFoundError} or \texttt{ModuleNotFoundError} at the Coder stage.
Constraint exhaustion occurs when the agentic retry budget is depleted without resolving upstream planning errors, causing the experiment to be marked as failed despite producing syntactically valid code.
API parameter mismatches the only recoverable category, stemming from the use of deprecated or incorrect Keras/TensorFlow arguments, which the Coder's internal self-correction mechanism can address within the iteration budget.
 
\paragraph{Distribution and Impact.}
The logs reveal that cross-modal hallucination and logical implementation drift constitute the dominant failure modes, exhibiting the largest downstream blast radius within the pipeline. While Tiny Scientist achieves a 100\% success rate on simple text classification tasks (such as GRU on IMDB), it collapses entirely when the Thinker introduces cross-modal
complexity. This pattern exposes a cascading error effect: an upstream reasoning failure in the Thinker forces the Coder into a state of soft deviation, iteratively attempting to reconcile an impossible plan until the hardware backend or iteration budget is exhausted. Unlike recoverable API mismatches, these reasoning-execution gaps cannot be resolved by self-correction alone and require a bidirectional verification loop between planning and execution stages. Taken together, these findings suggest that the primary bottleneck in current multi-agent research pipelines is not low-level code synthesis but high-level cross-modal reasoning and plan-execution consistency.

% \section{Test Examples}

\end{document}